\documentclass[journal]{IEEEtran}

\usepackage[fleqn]{amsmath}
\usepackage{amssymb}
\usepackage{amsfonts}
\usepackage{amsthm}
\usepackage{bm}

\newtheorem{corollary}{Corollary}
\newtheorem{proposition}{Proposition}

\usepackage{array}
\usepackage{booktabs}
\usepackage{multirow}
\usepackage{multicol}
\usepackage{tabularx}

\usepackage{graphicx}
\usepackage[caption=false,font=normalsize,
            labelfont=sf,textfont=sf]{subfig}

\usepackage{algorithmic}
\usepackage[ruled,linesnumbered]{algorithm2e}

\usepackage{xcolor}
\usepackage{soul}
\usepackage{textcomp}
\sethlcolor{yellow}

\usepackage{stfloats}
\usepackage{balance}

\usepackage{url}
\usepackage{hyperref}

\usepackage{xr}
\usepackage{xr-hyper}

\usepackage{verbatim}

\makeatletter

\newcommand*{\addFileDependency}[1]{%
    \typeout{(#1)}%
    \@addtofilelist{#1}%
    \IfFileExists{#1}{}{%
        \typeout{No file #1.}%
    }%
}

\makeatother

\newcommand*{\myexternaldocument}[1]{%
    \externaldocument{#1}%
    \addFileDependency{#1.tex}%
    \addFileDependency{#1.aux}%
}

\myexternaldocument{supplymentary}

\makeatletter

\newcommand\fs@norules{%
    \def\@fs@cfont{\bfseries}%
    \let\@fs@capt\floatc@ruled
    \def\@fs@pre{}%
    \def\@fs@post{}%
    \def\@fs@mid{\kern3pt}%
    \let\@fs@iftopcapt\iftrue
}

\makeatother

\def\BibTeX{%
    {\rm B\kern-.05em
    \sc i\kern-.025em b\kern-.08em
    T\kern-.1667em
    \lower.7ex\hbox{E}\kern-.125emX}%
}
\begin{document}


\title{MALNISI-Infused Graph Construction and Weighted Contrastive Loss based GAT for Unsupervised Representation Learning in HSI}

\author{Soumi Sarkar,~\IEEEmembership{Student Member, IEEE},
Subhadip Boral,
Sarbani Palit,~\IEEEmembership{Member, IEEE},
and Ashish Ghosh,~\IEEEmembership{Senior Member, IEEE}%

{\thanks{{Soumi Sarkar is a Ph.D. student of the Machine Intelligence Unit, Indian Statistical Institute, Kolkata, West Bengal 700108, India (e-mail: soumi.s}\_{r@isical.ac.in).}}%
\thanks{{Subhadip Boral is an Assistant Professor at Siksha 'O' Anusandhan, Bhubaneswar, Odisha 751030, India (e-mail: subhadipboral@soa.ac.in).}}%
\thanks{{Sarbani Palit is a Professor of the Computer Vision and Pattern Recognition Unit, Indian Statistical Institute, Kolkata, West Bengal 700108, India (e-mail: sarbanip@isical.ac.in).}}%
\thanks{{Ashish Ghosh is the Director of the International Institute of Information Technology, Bhubaneswar, Odisha 751003, India, and a Professor of the Machine Intelligence Unit, Indian Statistical Institute, Kolkata, West Bengal 700108, India (e-mail: ashish@iiit-bh.ac.in; ash@isical.ac.in).}}}%
}

\markboth{IEEE Transactions on Geoscience and Remote Sensing, 2026}{S. Sarkar, S. Boral, \MakeLowercase{\textit{(et al.)}: HSI Clustering}}

\maketitle

\begin{abstract}
Hyperspectral images (HSI) possess discriminating spectral characteristics with respect to land utilization due to high-dimensional information. Recently, unsupervised representation learning using graph neural networks has unveiled these utilization-specific complex spectral relations. This article has three main contributions. First, it proposes an index, the Mean Absolute Local Neighborhood Inverse Spectral Index (MALNISI), which measures the similarity between two pixels based on their neighborhood information. The index is used to construct a weighted adjacency matrix, giving the HSI a graph structure. Second, it proposes an upgraded Graph Attention Network that learns compact latent representations of hyperspectral data from the constructed graph. In this network, an enhanced multi-head attention mechanism incorporates the MALNISI weights as an auxiliary feature to emphasize learned attention between neighboring pixels. To alleviate over-smoothing and preserve fine-grained spectral details, the architecture integrates residual connections, skip projections, and feature normalization, which together enhance representation capability, stabilize learning, and accelerate convergence. Third, it proposes a weighted contrastive loss function that incorporates neighborhood-aware MALNISI between connected pixels, encouraging strongly connected pixels in the high-dimensional space to remain close in the latent space, and weakly connected pixels to stay apart. The resulting discriminative low-dimensional embeddings are then clustered using spectral clustering. Experimental analysis on six benchmark HSI datasets demonstrates the efficiency and necessity of the three aforementioned contributions, and the comparative study highlights superior performance over fourteen state-of-the-art approaches. The implementation of the proposed method is available in \href{https://github.com/soumisarkar02/MALNISI_GAT.git}{GitHub}.
\end{abstract}


\section{Introduction} \label{Introduction}
\IEEEPARstart{H}{yperspectral} images (HSI) \cite{chang2003hyperspectral} have rich spectral information, which enables accurate discrimination of different land-use and land-cover (LULC) materials. This discriminating nature must be learnt by unveiling complex inter-spectral correlations among the pixels. Unsupervised segmentation of LULC \cite{chang2003hyperspectral} focuses on partitioning the HSI into meaningful groups of pixels and addresses the combination of high-dimensional data with complex inter-spectral correlations. For unsupervised segmentation, graph-based learning methods, Graph Neural Networks (GNNs), have emerged as powerful tools in recent years for hyperspectral image analysis \cite{zhang2025elastic}. By representing pixels as nodes and similarity relationships as edges, graphs provide a natural framework for modelling the intrinsic structure of hyperspectral data \cite{zhang2025elastic}. In particular, graphs used in GNNs \cite{corso2024graph, wang2019scalable} effectively capture both local and non-local dependencies—unlike feature maps in CNNs \cite{SARKAR2026123258}, which are limited by fixed local receptive fields. This nature of learning is crucial for HSI data, where spectrally similar pixels may be spatially distant, and by leveraging graph connectivity, GNNs can effectively propagate information among such spectrally similar yet spatially distant pixels, enabling more robust representation learning for hyperspectral analysis.  

While GNNs are powerful in the context of HSI analysis, their effectiveness largely depends on the initial graph construction. Simple Euclidean-based adjacency fails to capture the non-linear manifold of hyperspectral data \cite{polowczykgraph}. Traditional spectral measures such as Spectral Angle Mapper (SAM) \cite{boardman1993spectral} and Spectral Information Divergence (SID) \cite{chang2000information} rely on pointwise comparisons, ignoring spatial context and remaining sensitive to noise and intra-class variability, often producing salt-and-pepper artifacts. Spatial–spectral methods like the Multiscale Spatial-Spectral Preprocessing Framework (MSSPF) \cite{akyurek2024spectral} incorporate local context but are computationally expensive due to patch-wise comparisons, while graph-regularization approaches using mutual nearest neighbor (MNN) \cite{cariou2020improving} require memory-intensive $k$-NN graphs. Anchor-based learning methods \cite{zhang2024bipartite,jiang2024structured} are also highly sensitive to initial anchor selection and often highly sensitive to noise. These limitations highlight the need for graph construction strategies that can efficiently model local and global spatial relationships while preserving fine-grained spectral characteristics. Graph construction \cite{zhang2025elastic} primarily depends on predefined similarity measures and superpixel segmentation, which may not fully capture fine-grained spectral differences between neighboring pixels, especially in regions with high spectral variability.

\begin{figure}[t]
\centering
\includegraphics[width=\linewidth]{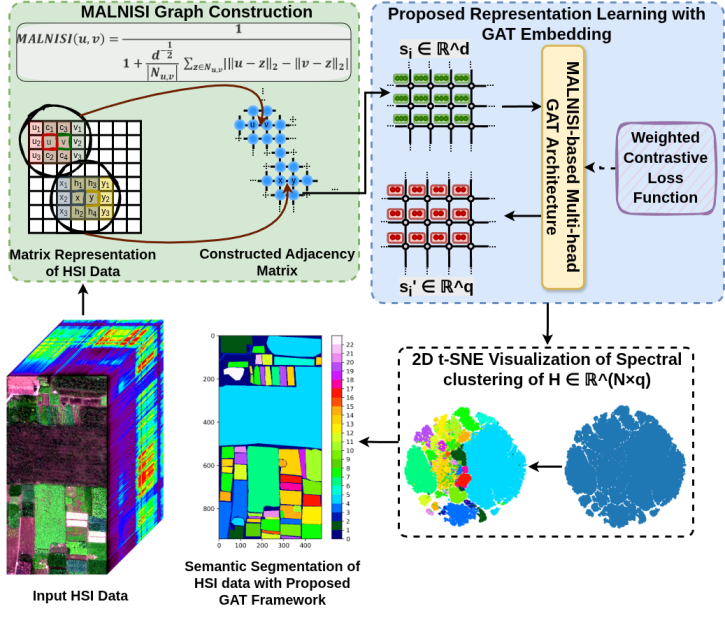}
\caption{Architectural overview of the proposed MALNISI-GAT framework, illustrating the three-stage semantic segmentation procedure: (1) MALNISI-based weighted graph construction, (2) unsupervised representation learning via multi-head Graph Attention Network with weighted contrastive loss, and (3) spectral clustering for segmentation.}
\label{fig:BD}
\end{figure}


In this article, a Mean Absolute Local Neighborhood Inverse Spectral Index (MALNISI) is proposed as the first contribution, focusing on the structural consistency between a pixel pair relative to a shared, localized neighborhood to capture contextual consistency-based spectral similarity rather than point-to-point spectral similarity. The absolute differences in spectral distance of both the points are taken with respect to every point in the common neighborhood, and then the inverse average score represents the proposed similarity index. The inverse function helps in bounding the index in $(0,1]$. The index rewards pairs that maintain a stable spectral relationship with their surroundings by evaluating how two pixels see their shared neighborhood.

The proposed index considers the union of individual neighborhoods which becomes more robust to noise, illumination variability, and minor spectral distortions that often affect individual pixel values. Here, neighborhood-level aggregation acts as a form of spatial smoothing, reducing the influence of random noise, while the use of relative comparisons within the neighborhood mitigates the effect of global illumination changes. Moreover, the use of absolute differences further enhances robustness by reducing the influence of local spectral distortions. The emphasis on local spectral consistency helps preserve structural patterns such as edges and textures, with tolerance for spatial misalignments. 

The MALNISI index is well-suited for constructing a weighted graph for HSI with $4$-neighborhood pixels as adjacent nodes inspired by graph-regularization approaches \cite{cariou2020improving}. Additionally, the pixels from the same LULC maintain consistent spectral properties with their neighbors, but at boundaries, this consistency breaks, producing weaker similarity. MALNISI reflects this trait and generates context-aware weighted graphs, leading to better preservation of LULC structures, which helps in improving clustering performance.

{Graph Attention Networks (GATs)} \cite{velivckovic2017graph} {among GNN architectures, are well-suited for lower-dimensional representation of graph nodes, as they can adaptively assign attention weights to neighboring nodes and selectively emphasize relationships between them. The second contribution of the work lies in the proposed MALNISI-guided upgraded GAT architecture, which exploits weighted graph information during attention computation, unlike conventional GATs. For this, the MALNISI edge weights are incorporated as auxiliary features to prevent their dilution, and, secondly, during attention score learning, they reweight the attention coefficients to guide learning. In addition, the proposed architecture employs multi-head attention to capture diverse neighborhood interactions and incorporates residual connections, skip projections, and feature normalization to alleviate over-smoothing and stabilize deep graph representation learning.}

{Unlike conventional methods that reconstruct only binary adjacency, the proposed weighted contrastive loss utilizes MALNISI-derived adjacency weights to preserve the relative manifold structure, ensuring that the pairwise relationships encoded in the high-dimensional space are faithfully retained in the low-dimensional embedding. This third contribution encourages strongly connected node pairs to remain close in the latent space while pushing negative pairs apart, leading to more discriminative low-dimensional representations. Besides, incorporating MALNISI weights into the positive-pair similarity term introduces neighborhood awareness absent in conventional contrastive loss.}

{Following dimensionality reduction, the learned latent embeddings from the MALNISI-GAT network are clustered to yield the final segmentation map. This is performed using spectral clustering (SC)} \cite{ng2001spectral}{, which has consistently shown strong performance for HSI analysis }\cite{zhang2025elastic}. 
{Thus, the main contributions of this work can be summarized as follows:}
\begin{itemize}
 \item {A similarity index, MALNISI $\in (0,1]$ that captures spectral similarity between two points considering common neighborhoods to put emphasis on local consistency.}
 \item {A MALNISI-guided Graph Attention Network (GAT) for robust nonlinear dimensionality reduction with upgraded attention mechanism to incorporate MALNISI as an auxiliary feature to prevent the loss of edge weights.}
 \item {A novel weighted contrastive loss that integrates cosine similarity with MALNISI score for incorporating neighborhood awareness and better latent space separability.}
\end{itemize}

The rest of the article is organized as follows. Section \ref{RW} reviews literature on hyperspectral clustering with graph-based learning. Section \ref{PI_MALNISI} and \ref{PM} present the proposed MALNISI index and the proposed GAT architecture. Section \ref{ER} reports experimental performance and comparative study, and conclusive remarks are given in Section \ref{CFW}.

\section{Related Literature}\label{RW}
\begin{table*}[hbp]
\centering
\caption{Taxonomy of SOTA HSI unsupervised segmentation methods.}
\label{tab:related_works}
\small
\resizebox{\textwidth}{!}{%
\begin{tabular}{ll p{4.5cm} p{4cm} p{4cm}}
\toprule
\textbf{Category} & \textbf{Methods} & \textbf{Mechanism} & \textbf{Key Advantages} & \textbf{Limitations} \\ \midrule
Linear Reduction & PCA, LDA & Linear projection to reduce spectral redundancy. & Computationally simple; fast execution. & Fails to capture non-linear structures. \\ \addlinespace
Centroid-Based & $K$-means, FCM & Iterative refinement of cluster centers. & Intuitive; incoorporates explainability. & Assumes spherical cluster shapes. \\ \addlinespace
Density-Based & DBSCAN, CFSFDP & Identification of clusters as high-density regions. & Finds arbitrary shapes; robust to noise. & Struggles with the curse of dimensionality. \\ \addlinespace
Graph-Based & \parbox[t]{2.8cm}{FSCAG, SGCNR, BGPC, SAPC} & Pixel-to-anchor adjacency and bipartite graph projection. & Near-linear scalability; captures connectivity. & Quality depends on anchor selection/kernels. \\ \addlinespace
Subspace-Based & \parbox[t]{2.8cm}{EnSC, SSC-BP, HESSC, SGLSC, EGFSC} & Representation as a union of low-dimensional subspaces. & Handles spectral variability and spatial edges. & Sensitive to manual graph design and noise. \\ \addlinespace
Deep Learning & \parbox[t]{2.8cm}{DCN, DSCRLE, DSCFS} & End-to-end non-linear feature extraction via AEs/CNNs. & High discriminative power; non-linear modeling. & Often suffers from fragmented feature integration. \\ \addlinespace
Hybrid Frameworks & \parbox[t]{2.8cm}{S$^3$AGC, NCSC} & Integration of contrastive loss or graph fusion in deep networks. & Synchronizes spatial-spectral features in latent space. & Potential sampling bias and training complexity. \\ \bottomrule
\end{tabular}%
}
\end{table*}

In earlier HSI clustering, Principal Component Analysis (PCA) \cite{datta2017unsupervised} and Linear Discriminant Analysis (LDA) \cite{murthy2017noisy} were used for lower-dimensional representation prior to partitioning. Manifold learning \cite{BORAL2023109447} introduced non-linear HSI data handling with noise sensitivity and high costs. When pixels were treated as separate entities during partitioning, Fuzzy $C$-Means (FCM) \cite{ahn1999delineation} and $K$-means \cite{zhou2022study} were unable to handle high spectral variability and non-spherical distributions. Density-Based Spatial Clustering of Applications with Noise (DBSCAN) \cite{backlund2011density} and Clustering by Fast Search and Finding of Density Peaks (CFSFDP) \cite{rodriguez2014clustering} identified arbitrary shapes but struggled with the curse of dimensionality. Graph-based frameworks like the Bipartite Graph-based Projected Clustering (BGPC) \cite{zhang2024bipartite} and Structured Anchor Projected Clustering (SAPC) \cite{jiang2024structured} address these limitations with non-neighboring pixel connectivity, anchor-pixel relationships, and Laplacian rank constraints. Fast Spectral Clustering with Anchor Graph (FSCAG) \cite{wang2017fast} and Scalable Graph-based Clustering with Nonnegative Relaxation (SGCNR) \cite{wang2019scalable} also address these by employing sparse anchor graphs and nonnegative relaxation to maintain linear complexity.

Subspace clustering improves segmentation by assuming pixels of the same class lie within low-dimensional subspaces. Sparse Subspace Clustering-BP (SSC-BP) \cite{elhamifar2013sparse}, Elastic Net Subspace Clustering (EnSC) \cite{you2016oracle} and Hierarchical Sparse Subspace Clustering (HESSC) \cite{rafiezadeh2020hierarchical} leverage self-expressiveness to reconstruct target pixels from a sparse linear combination of the HSI dictionary. Superpixel-level Global and Local Similarity-based Clustering (SGLSC) \cite{zhao2021superpixel} and Elastic Graph Fusion Subspace Clustering (EGFSC) \cite{zhang2025elastic} utilize a dual-graph fusion strategy to merge spatial superpixel and spectral band relationships into a unified latent representation.

Modern advancements like Non-local Contrastive Subspace Clustering (NCSC) \cite{cai2022superpixel} introduce deep subspace clustering via a superpixel pooling autoencoder and neighborhood contrastive regularization, which attract spatial neighbors while repelling dissimilar samples in a contracted latent space. Deep Spectral Clustering Feature Selection (DSCFS) \cite{zhao2025deep} also combines deep learning-based HSI representation with adaptive feature projection and extracts key features. Concurrently, deep representation learning \cite{ghosh2026data} based methods like Deep Clustering Network (DCN) \cite{zhai2021hyperspectral} utilize Autoencoders (AE) \cite{chakraborty2019integration} and Convolutional Neural Networks (CNN) \cite{hsu2017cnn} to capture intricate nonlinearities, while their success relies on the integration of feature-learning backbones and clustering objectives.

Recent research has pivoted toward hybrid frameworks to synthesize these disparate strengths. Deep Spectral Clustering with Regularized Linear Embedding (DSCRLE) \cite{zhao2023deep} combines regularized linear embedding from deep networks with a graph Laplacian matrix obtained from anchor graph construction and solves out-of-sample problems. Scalability and noise robustness are emphasized in Spectral-Spatial Sparse Anchor Graph Clustering (S$^3$AGC) \cite{chen2023spectral}, which employs superpixel-based denoising and band-based graph convolutions. Despite these advancements, existing frameworks often rely on static kernels, suffer from sampling bias, or push apart pixels that are spectrally similar but not spatially adjacent. To address these limitations, the proposed methodology uses a GAT for attention-based relationship learning and transitive feature aggregation via multi-hop message passing. By integrating the MALNISI index for noise-robust spectral guidance and keeping the same notion in latent space through a weighted contrastive loss function, the framework ensures improved performance. A comprehensive summary of state-of-the-art (SOTA) and contrasting improvements are presented in Tables \ref{tab:related_works} and \ref{tab:proposed_vs_related}.

\section{Proposed Index: Mean Absolute Local Neighborhood Inverse Spectral Index (MALNISI)} \label{PI_MALNISI}


\textbf{A. Mathematical Formulation: }
\noindent Let $u, v \in \mathbb{R}^d$ denote two points and their joint neighborhood be defined as:
\begin{equation}
\mathcal{N}_{u,v} = \left( \mathcal{N}(u) \cup \mathcal{N}(v) \right) \setminus \{u, v\}.
\end{equation}
Here, $\mathcal{N}(\cdot)$ denotes the neighborhood function and $\setminus$ indicates set subtraction. 

To quantify the difference between $u$ and $v$ with respect to $\mathcal{N}_{u,v}$, the proposed Spectral Index ($SI$) is defined as: 
\begin{equation} \label{eq:SDD}
SI(u,v) = \sum_{z \in \mathcal{N}_{u,v}} \big| \|u-z\|_2 - \|v-z\|_2 \big|. 
\end{equation}
where $z$ is a point in the shared neighborhood $\mathcal{N}_{u,v}$.


While $SI$ provides a raw measure of difference, it is unbounded and varies with the neighborhood function and dimensions. To transform this into a non-negative and symmetric similarity index, $MALNISI$, which obeys the identity property and is bounded in $(0, 1]$, is defined as:
\begin{equation}
\label{eq:short_madd}
MALNISI(u,v) = \frac{1}{1 + \frac{d^{-1/2}}{\lvert \mathcal{N}_{u,v} \rvert} SI(u,v)}.
\end{equation}

\begin{table}[htbp]
\centering
\caption{Summary of technical challenges in SOTA and proposed solutions in the MALNISI-GAT framework.}
\label{tab:proposed_vs_related}
\resizebox{\linewidth}{!}{%
\begin{tabular}{p{2.5cm}|p{6cm}}
\toprule
\textbf{Challenges in SOTA} & \textbf{Proposed Solution (MALNISI-GAT Network)} \\ \midrule \midrule
Spectral Noise Interference & Integrates noise-robust MALNISI in initial graph construction. \\ \midrule
Spatially Distributed Spectrally Similar Regions & Graph networks have transitive feature aggregation across multiple layers that strongly pull same-category pixels. \\ \midrule
Complex \& Non-Compact Data & Weighted contrastive loss forces to maintain spectrally consistency while embedding. \\ \midrule
\end{tabular}}
\end{table}

\noindent The \textbf{dimensionality normalization factor} $\boldsymbol{d^{-1/2}}$ transforms cumulative spectral distance into a band-averaged index and makes $MALNISI$ consistently sensitive and discriminative in higher dimensions. Without normalization,  $SI(u,v)$ grows proportionally to $d^{1/2}$ as $\lVert \cdot \rVert_2$ scales monotonically with $d$ and makes $MALNISI$ to saturate towards zero in high dimensions. On the other hand, the \textbf{normalization factor} $\boldsymbol{1/|\mathcal{N}_{u,v}|}$ makes $MALNISI$ unbiased towards the neighborhood function.



\textbf{B. Theoretical Properties of MALNISI: }
The properties of MALNISI are stated in this section, and the proofs and discussion are presented in Section \ref{app:proof} of the supplementary documents.
\begin{proposition} \label{prop: NNP}
$MALNISI(u,v)$ satisfies the non-negativity property, $MALNISI(u,v) > 0, \quad \forall u, v \in \mathbb{R}^d$.
\end{proposition}

\begin{proposition}
The index $MALNISI(u,v)$ satisfies the symmetry property, $MALNISI(u,v) = MALNISI(v,u)$.
\end{proposition}

\begin{proposition}
The $MALNISI(u,v)$ holds the identity property, i.e., if $u = v$, then $MALNISI(u,v) = 1$.
\end{proposition}

\begin{proposition} \label{prop:TI}
$SI(u,v)$ does not satisfy the triangle inequality property.
\end{proposition}

\begin{corollary}
$MALNISI(u,v)$ does not satisfy the triangle inequality.
\end{corollary}

\begin{proposition} \label{prop:IIP}
The index $MALNISI(u,v)$ does not strictly hold the identity of indiscernibles property.
\end{proposition}

\begin{proposition}
The $SI(u,v)$ is unbounded over $\mathbb{R}^d \times \mathbb{R}^d$.
\end{proposition}

\begin{proposition} \label{prop:sim_dissim}
The $MALNISI(u,v)$ is bounded in the range $(0, 1]$ and $MALNISI(u,v)$ satisfies the following properties:
\begin{enumerate}
 \item \underline{\textit{Spectral Similarity:}} For nearly similar points, the $MALNISI(u,v) = 1$ as $\|u - v\|_2 \to 0$.
 \item \underline{\textit{Spectral Dissimilarity:}} For dissimilar points, when $u$ and $v$ are spectrally very very far, i.e., without the loss of generality, $\|u - v\|_2$ approaches its maximum attainable value, then $MALNISI(u,v)$ approaches its minimum attainable value $0$.
\end{enumerate}
\end{proposition}
These properties show that MALNISI does not qualify as a metric but qualify as an index.

\textbf{C. Generalizability of the Neighborhood Function: }
The definition of the neighborhood function $\mathcal{N(.)}$ in the $MALNISI$ framework can be adapted to suit specific application domains. 
\begin{enumerate}
\item \underline{\textit{Invariance of Theoretical Properties: }}
A significant advantage of the $MALNISI$ index is that its fundamental theoretical properties are invariant to the specific selection of the neighborhood set; the Proposition \ref{prop: NNP} -- Proposition \ref{prop:sim_dissim} will still hold.
\item \underline{\textit{Global Neighborhood and the Triangle Inequality: }}
An important theoretical distinction arises when the neighborhood function yields a common neighborhood for all data points (i.e., when the entire dataset is utilized as the neighborhood for every point). When there is a fixed neighborhood for all pairs $(u,v)$, $(v,w)$, and $(u,w)$, the triangle inequality is strictly satisfied:
\begin{equation}
SI(u,w) \le SI(u,v) + SI(v,w).
\end{equation}
In this context, the spectral index $SI$ ceases to be a localized dissimilarity measure and becomes a valid pseudo-metric across the entire spectral feature space. 
\end{enumerate}

\textbf{D. Adaptability to Noise \& Spectral Distortion: }
If noise affects $\mathcal{N}_{u,v}$ then subtracting each neighbor from $u$ and $v$ effectively cancels out the noise. This notion works similarly for spectral distortion, as feature-specific distortion will be present for each point and will be negated during the calculation via Equation \ref{eq:SDD}. 

\section{Proposed Methodology} \label{PM}
\textbf{A. Problem Formulation: }
Let $\mathbf{X} = \{\mathbf{x}_1, \mathbf{x}_2, \dots, \mathbf{x}_N\} \in \mathbb{R}^{N \times d}$ denote a HSI data, where $N = Row \times Col$ is the total number of pixels and $d$ is the number of spectral bands. 
Given $\mathbf{X}$, the goal of clustering is to partition the $N$ pixels into $K$ disjoint groups $\{C_1, C_2, \dots, C_K\}$, i.e., to learn a mapping function that assigns each pixel $\mathbf{x}_i$ to a cluster label based on its spectral signatures\footnote{Note: A single point in Section \ref{PI_MALNISI} corresponds to a single HSI pixel in Section \ref{PM}.}.

\textbf{B. Motivation \& Intuition: } 
Spectral Clustering (SC) is highly effective in HSI clustering, as it captures the complex, non-linear distribution of hyperspectral data \cite{zhang2025elastic}. However, SC is computationally expensive with $O(N^3)$, as pairwise similarity is calculated, and is also susceptible to the ``curse of dimensionality''. Due to this, formation of an ineffective adjacency matrix $\mathbf{A}$ leads to poor clustering and LULC boundary identification. To enhance the efficacy of SC, it is prudent to perform dimensionality reduction with structural separability. Feature selection approaches are less suitable here, as they lose information, and extraction methods \cite{BORAL2023109447} including PCA, Isomap, Locally Linear Embedding (LLE), and Uniform Manifold Approximation and Projection (UMAP), are not fully suitable as they are confined to non-linear structural preservation. On the other hand, GAT are capable of not only addressing these issues but also have spectral discrimination with a compact latent space projection \cite{11223040}. In GAT, though, adjacency is established only among neighboring nodes, but information from distant neighbors is propagated through message-passing mechanisms \cite{liu2025joint}. 

In HSI, pixels of a specific LULC generally form cohesive spatial groups rather than isolated, scattered patterns. This observation motivates the integration of spatial context into the proposed architecture. To achieve this, the proposed framework leverages both the advantage of the message passing in GAT and incorporates localized spatial context by utilizing the immediate neighbors of each pixel as the neighborhood function within the MALNISI computation. 

\textbf{C. MALNISI Graph Construction: }
The undirected weighted graph $\mathcal{G} = (\mathcal{S}, \mathbf{A})$ from $\mathbf{X}$ is constructed as follows.
\begin{itemize}
 \item $\mathcal{S} = \{ s_1, s_2, \dots, s_N \}$ is the set of d-dimensional nodes corresponding to respective pixels, where $ s_i = \mathbf{x}_i$,
 \item $\mathbf{A} \in \mathbb{R}^{N \times N}$ is the weighted adjacency matrix.
\end{itemize}
The edge weight between nodes $s_i$ and $s_j$ is set as,
\begin{equation}
A_{ij} =
\begin{cases}
MALNISI(s_i, s_j), & \text{if } s_j \in \mathcal{M}(s_i), \\
0, & \text{otherwise},
\end{cases}
\end{equation}
where $\mathcal{M}(s_i)$ denotes the $N_4(s_i)$ obtained from $\mathbf{X}$ and $\mathcal{N}(\cdot)$ in MALNISI is $N_8(\cdot)$.

\begin{figure}[t]
\centering
\includegraphics[width=0.99\linewidth]{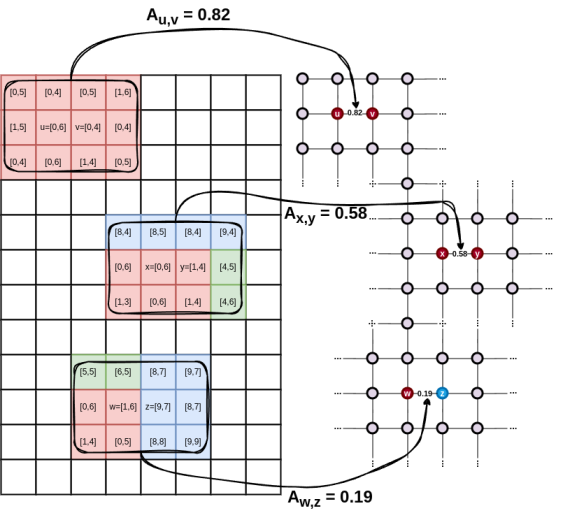}
\caption{Illustration and analysis of the MALNISI-based graph construction, where different colors represent different pixel category.} 
\label{fig:GC}
\end{figure}

This allows the subsequent GAT layers to focus more on spectrally similar neighbors and ignore dissimilar ones, even if they are spatially adjacent. This process of transforming the spatial grid of an HSI into a weighted adjacency graph is illustrated in Figure \ref{fig:GC}. We discuss the following case studies:

\textbf{Case 1 - Pixels and neighbors from the same LULC (}$\mathbf{A_{u,v} = 0.82}$\textbf{):} This represents a scenario where not only pixels $u$ and $v$ belong to the same category, but also their surrounding neighborhood also belongs to the same category. Because the spectral signatures are consistent throughout the local window, MALNISI produces a high weight ($0.82$), indicating a very strong connection between $u$ and $v$.

\textbf{Case 2 - Pixels from the same LULC and neighbors from different LULC (}$\mathbf{A_{x,y} = 0.58}$\textbf{):} In this case, pixels $x$ and $y$ still belong to the same category, but their neighborhood is heterogeneous, containing pixels from other categories (represented by the mix of red, blue, and green blocks). The presence of these neighbors from different categories causes the MALNISI value to decrease with the growing number of pixels from other categories.

\textbf{Case 3 - Pixels and neighbors from different LULC (}$\mathbf{A_{w,z} = 0.19}$\textbf{):} This case occurs when pixels $w$ and $z$ belong to different categories, i.e., part of an edge or boundary (the red and blue regions). The presence of pixels from different categories in the common neighborhood creates high spectral variations and low MALNISI. It effectively signals the GAT to place less emphasis on these neighbors.

The computational complexity of the graph construction is $\mathcal{O}(N*|\mathcal{N}(.)|* d) \approx \mathcal{O}(N*d)$ as $|\mathcal{N}(.)| << d$.


\textbf{D. GAT-Based Adjacency Modeling with Dimension Reduction: }
After the graph construction, the objective is to learn a compact latent representation 
$\mathbf{H} \in \mathbb{R}^{N \times q}$ with $q \ll d$ through a mapping
\begin{equation}
\Phi_{\Theta} : (\mathbf{X}; \mathbf{A}) \rightarrow (\mathbf{H}; \hat{\mathbf{A}}),
\end{equation}
such that the embedding $\mathbf{H}$ preserves the intrinsic spectral relationships encoded in $\mathbf{A}$ and $\Theta$ denotes trainable parameters. 

\begin{figure}[ht]
\centering
\includegraphics[width=\linewidth]{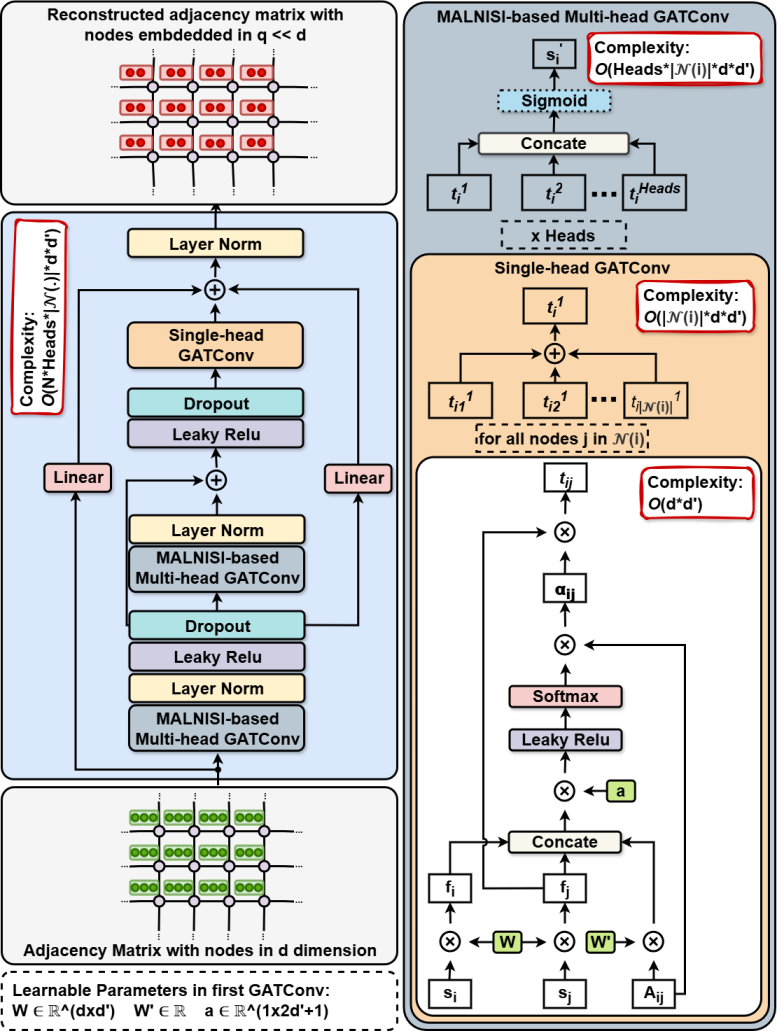}
\caption{Detailed illustration of the proposed MALNISI-GAT architecture.}
\label{fig:MA}
\end{figure}

\textit{\textbf{1) MALNISI-based Upgraded Graph Attention Network: }}
The left column of Figure \ref{fig:MA} illustrates the proposed GAT architecture, and the multi-head attention mechanism is illustrated in the right column of Figure \ref{fig:MA}, highlighting how MALNISI guides the learning of attention weights. 

The primary objective of this framework is to inject the relative importance of neighboring pixels for individual pixels in GAT learning. To achieve this, MALNISI-based weights are incorporated in two ways: first, as feature attributes to enrich the input representation; and second, to re-emphasize the learned attention coefficients according to the MALNISI values. Multi-head attention is employed in the network to capture different views from the same neighborhood, thereby producing richer feature representations. However, in the final layer, single-head attention is utilized to satisfy the user-defined latent space dimensionality requirement.

Despite its effectiveness, repeated neighborhood aggregation during message passing in GAT may cause intrinsically dissimilar nodes to develop highly similar representations, leading to an over-smoothing effect. Layer normalization applied after each attention layer to rescale the node representations, which improves embedding compactness, cluster separability, and mitigates the effect. The Leaky ReLU activation function \cite{9209059} introduces non-linearity into the network, enabling it to learn complex relationships. Additionally, dropout serves as a regularization mechanism by randomly deactivating a subset of feature dimensions and attention coefficients during training, thereby reducing overfitting. Apart from this, residual connections via linear transformation layers help mitigate the over-smoothing problem by appending lower-level discriminating features. 


\textit{\textbf{2) Proposed Weighted Contrastive Loss: }}
GAT does not explicitly enforce discriminative representations in the embeddings, and therefore, to enforce discrimination, contrastive loss \cite{11223040} is incorporated. To achieve discrimination, if $u$ and $v$ are two pixels from the same region, then they are to be embedded very closely (high $A_{ij}$); and the separation grows when they belong to different regions (low $A_{ij}$). The following component of contrastive loss encourages this discrimination between dissimilar node embeddings and vice versa. 
\begin{figure}[ht]
\centering
\includegraphics[width=0.99\linewidth]{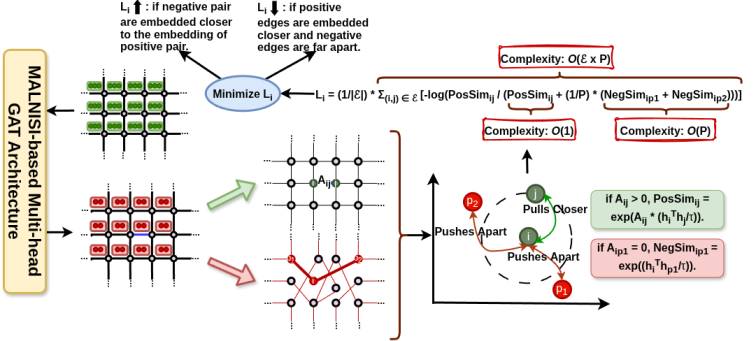}
\caption{Geometric visualization of the proposed weighted contrastive loss.}
\label{fig:Loss}
\end{figure}
For each positive edge ($\mathbf{A}_{ij} >  0$) the similarity score is:
\begin{equation}
r_{ij} =
\frac{\mathbf{h}_i^\top \mathbf{h}_j}{\tau},
\label{eq:con1}
\end{equation}
where $\mathbf{h}_i$ is the output from the $i$-th node in last layer of the proposed network and $\tau$ is a temperature parameter. However, this cosine similarity treats all connected node pairs uniformly, ignoring the varying spectral strengths of the edges, and is neighborhood-agnostic. To resolve this, the similarity score is modulated by the MALNISI weights, which contain the local neighborhood relationships that were captured during graph construction (shown in Figure \ref{fig:Loss}) and are defined as,
\begin{equation}
\tilde{r}_{ij} = A_{ij} \cdot r_{ij}.
\label{eq:con2}
\end{equation}
For each anchor node $s_i$, $P$ negative nodes ($\mathbf{A}_{ij} =  0$) are randomly sampled, and their similarity scores ($r_{ip}$) are defined with equation \ref{eq:con1}. $r_{ip}$ remain unweighted as they do not correspond to any existing edges, i.e., MALNISI edge weights do not exist. Thus, the weighted contrastive loss is,
\begin{equation}
\mathcal{L}_{i} = \frac{1}{|\mathcal{E}|} \sum_{(i,j) \in \mathcal{E}} -\log
\frac{
\exp(\tilde{r}_{ij})
}{
\exp(\tilde{r}_{ij})
+
\sum_{p=1}^{P}
\exp(r_{ip})
},
\label{eq:cl}
\end{equation}
where $\mathcal{E}$ represents the set of all positive edges from the anchor node $s_i$. Observed from Equation \ref{eq:cl}, when the neighborhood importance in the learned embedding deviates from that of the original feature space (i.e., high $\sum_{p=1}^{P} \exp(r_{ip})$ and low $\exp(\tilde{r}_{ij})$), highly similar node pairs contribute more to weighted cosine-distance penalties. Consequently, the loss increases, forcing the preservation of local structural relationships during representation learning, and it is expected to decrease during optimization. Figure \ref{fig:Loss} shows this and also highlights that the computational complexity of the contrastive loss for a single node is $\mathcal{O}(|\mathcal{E}|*P)$. Thus, for all $N$ nodes, computation complexity becomes $\mathcal{O}(N*|\mathcal{E}|*P)$.

Therefore, the GAT network dynamically generates the node embeddings $\mathbf{H}$ by adaptively learning neighbor importance, while the weighted contrastive loss acts as a structural regularizer that forces these embeddings to respect the topology of the hyperspectral graph. Once the network converges, $\mathbf{H}$ is fed into SC for semantic segmentation. Computational complexity of SC is $\mathcal{O}(N^3)$. 





\textbf{E. Computation Complexity: }
The model takes $\mathbf{X}$, $\mathbf{A}$ and edge weight as input and learns $\Phi_{\theta}$, shown in Figure \ref{fig:MA}. From the figure, the complexity of the first multi-head GATConv is $\mathcal{O}(N*Heads*|\mathcal{N}(.)|*d*d')$. Here, $Heads*|\mathcal{N}(.)| << d,N$, so, the complexity becomes $\mathcal{O}(N*d*d')$. Similarly, for the second multi-head GATConv is $\mathcal{O}(Nd'*d')$ and for the last multi-head GATConv is $\mathcal{O}(N*d'*q)$. Now, since $d>d'>q$, the overall complexity is $\mathcal{O}(N*d*d')$. The overall complexity of the proposed framework can be expressed as,
\begin{equation}
\mathcal{O}(N*d) + \mathcal{O}(N*d*d') + \mathcal{O}(N*|\mathcal{E}|*P) + \mathcal{O}(N^3).
\end{equation}
Since $|\mathcal{E}|$, $P$, $d$ and $q$ are typically much smaller than $N$, the proposed approach scales with cubic complexity.

\section{Experimental Setup \& Performance Analysis} \label{ER}
Experiments are conducted in a Red Hat Enterprise Linux Server 7.9 system with a 64-bit OS, equipped with an Intel® Xeon® Gold 5317 CPU @ 3.00 GHz (48 cores), 500 GB of RAM, and 13 TB of storage. In this experiment, the Adam optimizer with a learning rate $\eta = 0.01$ is used.

Table \ref{tab:datasets} demonstrates the detailed description of the dataset used for effectiveness evaluation of the proposed framework over diverse LULC categories and varying spatial resolutions. Table \ref{tab:metrics} provides details of the evaluation metric employed to quantitatively evaluate information-theoretic similarity, cluster compactness, and inter-cluster separability of the proposed work. To follow a standardized evaluation \cite{zhang2025elastic}, pixels categorized as unlabeled (designated with label $0$ in ground truth (GT) maps) are excluded after determination of $\mathbf{H}$.
\begin{table}[ht]
\centering
\caption{Description of the benchmark HSI datasets used for the performance evaluation.}
\resizebox{\linewidth}{!}{
\begin{tabular}{p{2.2cm}|p{1.1cm}|p{1.9cm}|ccc|p{1.5cm}}
\hline
\textbf{Dataset} & \textbf{Sensor} & \textbf{Location} & \textbf{Image Size} & \textbf{Bands} & \textbf{Categories} & \textbf{Data Type} \\
\hline
Indian Pines (IndianP) \cite{grana2017hyperspectral} & AVIRIS & Indiana, USA & $145 \times 145$ & 200 & 16 & Agricultural \\ \hline
Pavia University (PaviaU) \cite{grana2017hyperspectral} & ROSIS & Pavia, Italy & $610 \times 340$ & 103 & 9 & Urban \\ \hline
Salinas \cite{grana2017hyperspectral} & AVIRIS & Salinas Valley, CA, USA & $512 \times 217$ & 204 & 16 & Agricultural \\ \hline
HongHu \cite{zhong2020whu} & Nano-Hyperspec & Honghu, Hubei, China & $940 \times 475$ & 270 & 22 & Agricultural \\ \hline
Botswana \cite{grana2017hyperspectral} & Hyperion & Okavango Delta, Botswana & $1476 \times 256$ & 145 & 14 & Wetland / Vegetation \\ \hline
Kennedy Space Center (KSC) \cite{grana2017hyperspectral} & AVIRIS & Florida, USA & $512 \times 614$ & 176 & 13 & Wetland / Vegetation \\ \hline
\end{tabular}}
\label{tab:datasets}
\end{table}
\begin{table}[ht]
\centering
\caption{Metrics used for the performance evaluation.}
\resizebox{\linewidth}{!}{
\begin{tabular}{p{1cm}|p{2.5cm}|p{5cm}|p{1cm}|p{1.5cm}}
\hline
\textbf{Type}  & \textbf{Meric}  &\textbf{Purpose} & \textbf{Range} & \textbf{Optimal} \\ 
 & & & & \textbf{Direction} \\  \hline \hline
 & \textbf{Overall Accuracy (OA)} & Overall correct assignment of pixels to corresponding true group & $[0,100]$ & $\mathbf{\uparrow}$ \\ \cline{2-5}
 & \textbf{Average Accuracy (AA)} & Mean group-wise accuracy with dominant-group bias mitigation& $[0,100]$ & $\mathbf{\uparrow}$ \\ \cline{2-5}
 & \textbf{Normalized Mutual Information (NMI)} & Similarity between clustering assignments and true group from an information-theoretic perspective with by chance agreement & $[0,1]$  & $\mathbf{\uparrow}$ \\ \cline{2-5}
\textbf{External Index} & \textbf{Adjusted Rand Index (ARI)} & Pairwise similarity between clustering assignments and true group from an information-theoretic perspective & $[-1,1]$ & $\mathbf{\uparrow}$ \\ \cline{2-5}
 & \textbf{Adjusted Mutual Information (AMI)} & Clustering consistency and structural correspondence between predicted and true group with by chance agreement reduction & $[-\infty,1]$ & $\mathbf{\uparrow}$ \\ \cline{2-5}
 & \textbf{Fowlkes–Mallows Index (FMI)}  & Pairwise clustering consistency and structural correspondence between clusters and true labels & $[0,1]$  & $\mathbf{\uparrow}$ \\ \cline{2-5}
 & \textbf{Homogeneity} & Each cluster contains data from a single true group or not & $[0,1]$  & $\mathbf{\uparrow}$ \\ \cline{2-5}
 & \textbf{Completeness} & Measure whether all samples of a given category are assigned to the same cluster  & $[0,1]$  & $\mathbf{\uparrow} $\\ \cline{2-5}
 & \textbf{Kappa coefficient} & Agreement between predicted clusters and corresponding true group with mitigation of agreement by chance effect & $[-1,1]$ & $\mathbf{\uparrow}$ \\ \cline{2-5}
 & \textbf{Purity}  & Measures cluster homogeneity by measuring the dominant category consistency within each cluster & $[0,1]$  & $\mathbf{\uparrow}$ \\ \hline \hline
 & \textbf{Davies–Bouldin Index (DB)} & Evalutes the similarity between each cluster and most similar one & $[0,\infty]$ & $\mathbf{\downarrow}$  \\ \cline{2-5}
\textbf{Internal Index} & \textbf{Silhouette Score (SH)} & Combines intra- and inter-cluster distances  & $[-1,1]$ & $\mathbf{\uparrow}$ \\ \cline{2-5}
 & \textbf{Calinski–Harabasz Index (CH)} & Ratio of between to withincluster dispersion  & $[0,\infty]$ & $\mathbf{\uparrow}$ \\ \cline{2-5}
 & \textbf{Dunn Index} & Ratio of minimum inter-cluster distance to maximum intra-cluster diameter & $[0,\infty]$ & $\mathbf{\uparrow}$ \\
 \hline
\end{tabular}}
\label{tab:metrics}
\end{table} 

\textbf{A. Qualitative and quantitative performance analysis on benchmark data with visualizations: }
\begin{table}[htbp]
\centering
\caption{Qualitative improvement in discriminative embedding by the proposed framework and comparative spatial segmentation.}
\label{tab:visual_results}
\resizebox{!}{0.32\textheight}{
\begin{tabular}{p{3cm}|p{2.15cm}|p{2.15cm}|p{1.4cm}|p{2.4cm}}
\hline
\textbf{Dataset specific GT of semantic segmentation} 
& \multicolumn{3}{c|}{\textbf{2D t-SNE visualization of learned and clustered embeddings}} 
& \textbf{Semantic segmentation by proposed method} \\ \cline{2-4}
& \textbf{Raw+GT} 
& \textbf{GAT+GT} 
& \textbf{GAT+Predicted} 
& \\ \hline
\raisebox{1.7\height}{
\rotatebox[origin=c]{90}{\textbf{IndianP}}
\hspace{1mm}}
\includegraphics[width=0.12\textwidth]{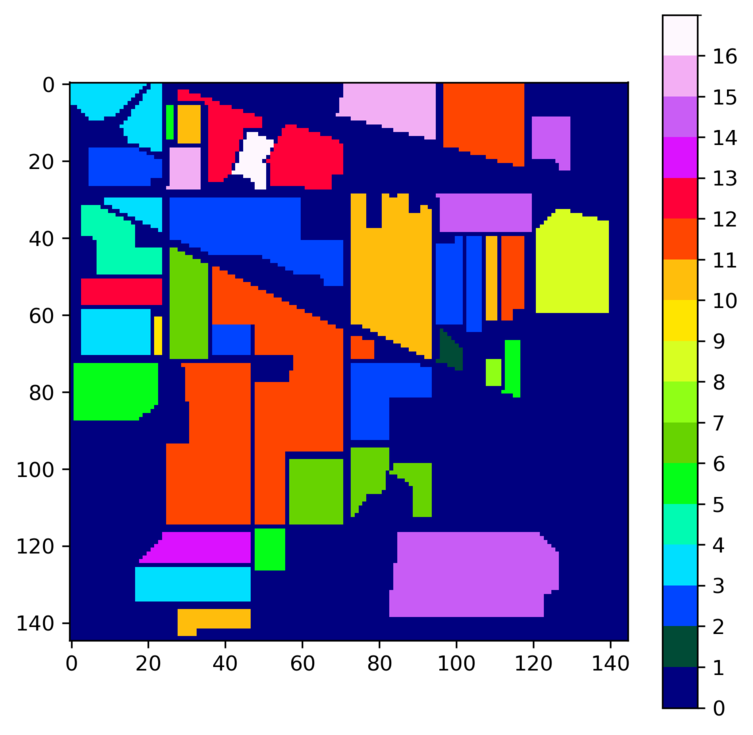}
&
\raisebox{0.05\height}{\includegraphics[width=0.12\textwidth]{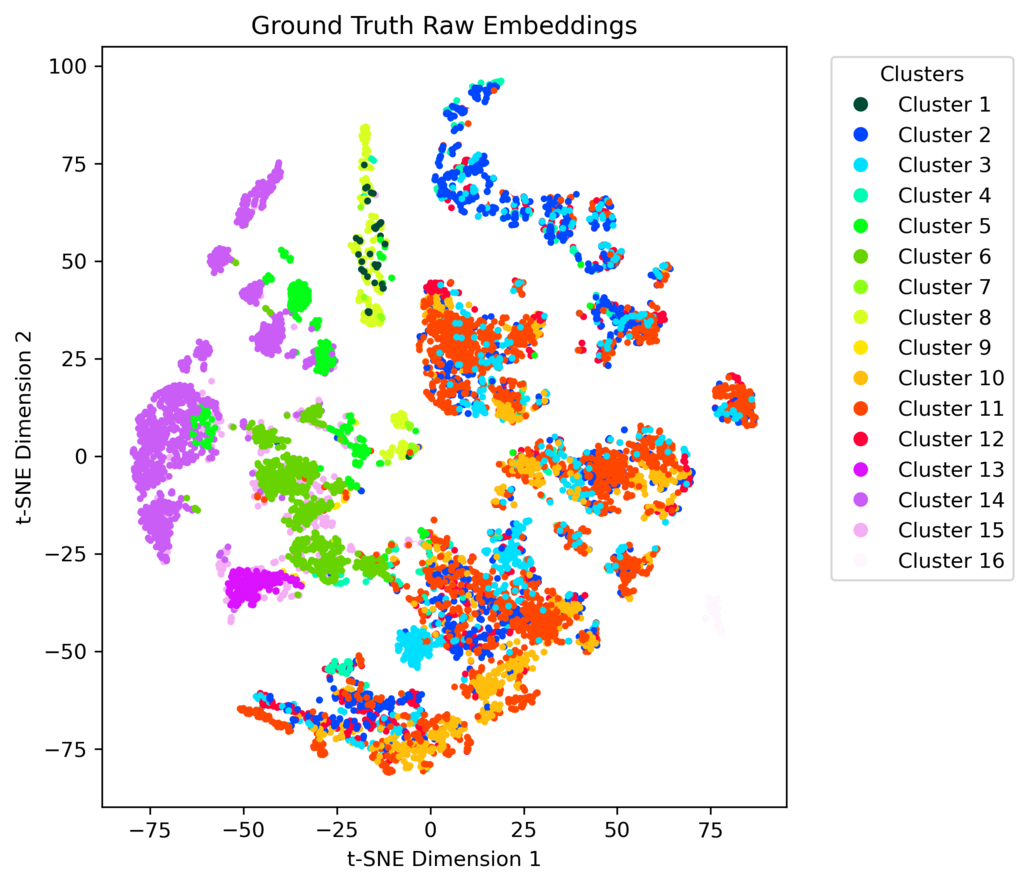}}
&
\raisebox{0.05\height}{\includegraphics[width=0.12\textwidth]{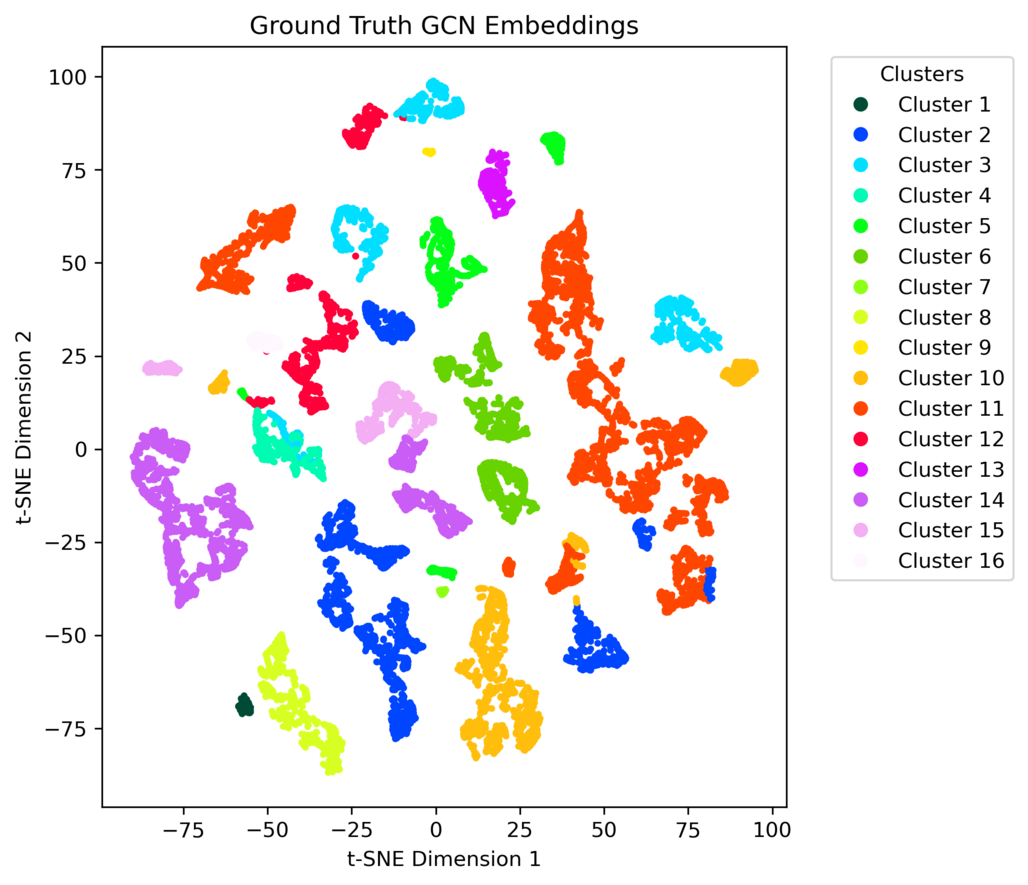}}
&
\raisebox{0.05\height}{\includegraphics[width=0.12\textwidth]{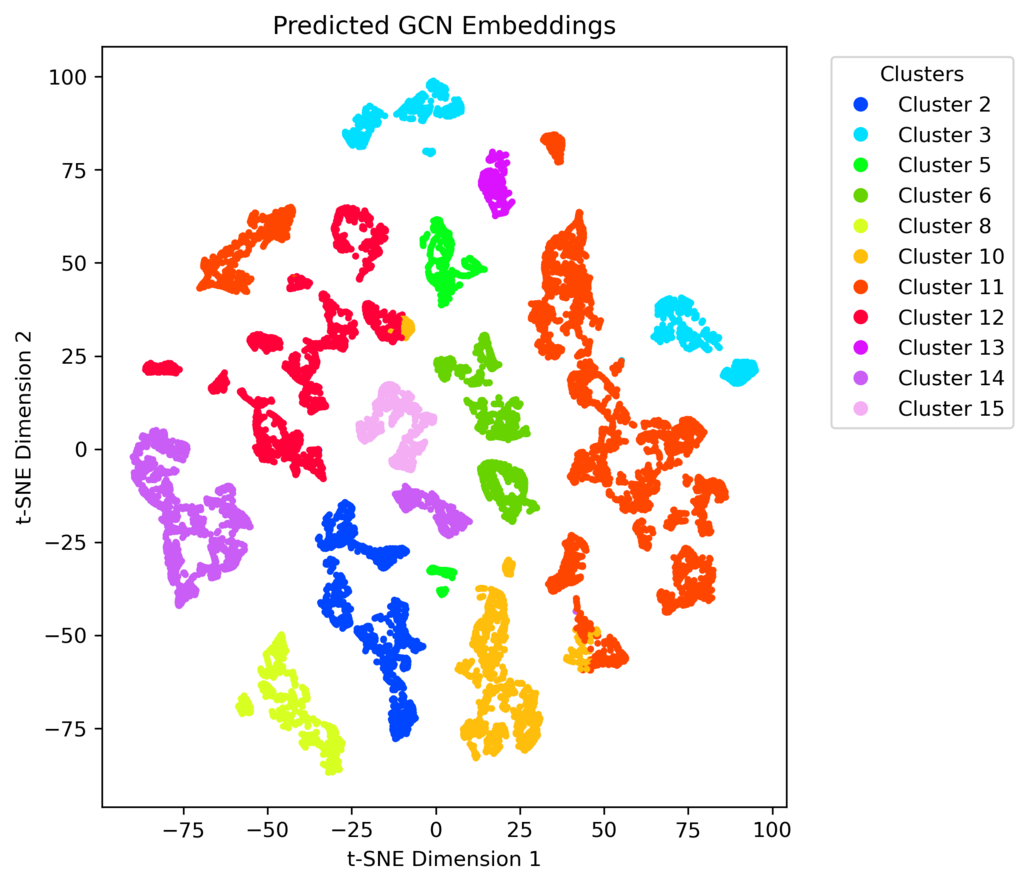}}
&
\includegraphics[width=0.12\textwidth]{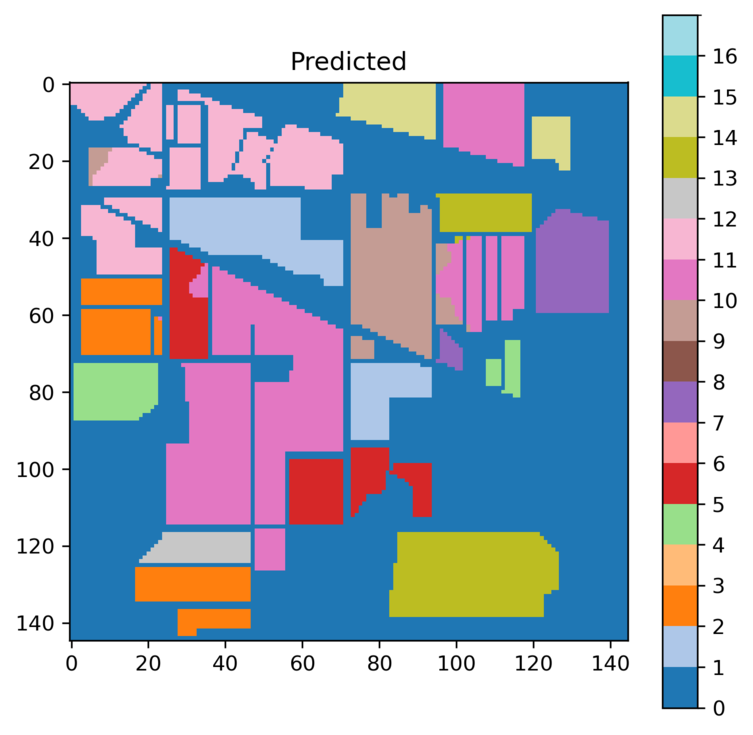}
\\ 
& \centering \textbf{DB:} 4.3652 & \centering \textbf{DB:} 1.8246 & \centering \textbf{DB:} 1.8208 & \\
& \centering \textbf{SH:} -0.0412 & \centering \textbf{SH:} 0.1242 & \centering \textbf{SH:} 0.206 & \\
\hline
\raisebox{2.6\height}{
\rotatebox[origin=c]{90}{\textbf{PaviaU}}
\hspace{1mm}}
\includegraphics[width=0.12\textwidth]{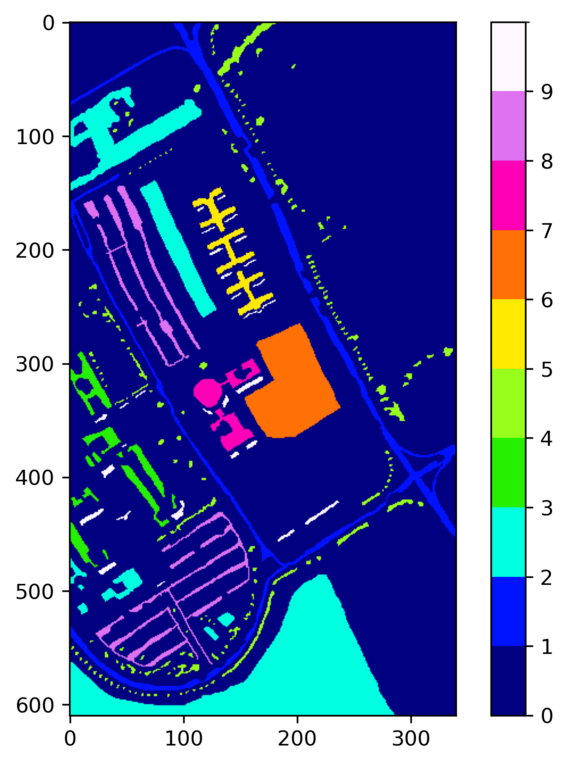}
&
\raisebox{0.25\height}{\includegraphics[width=0.12\textwidth]{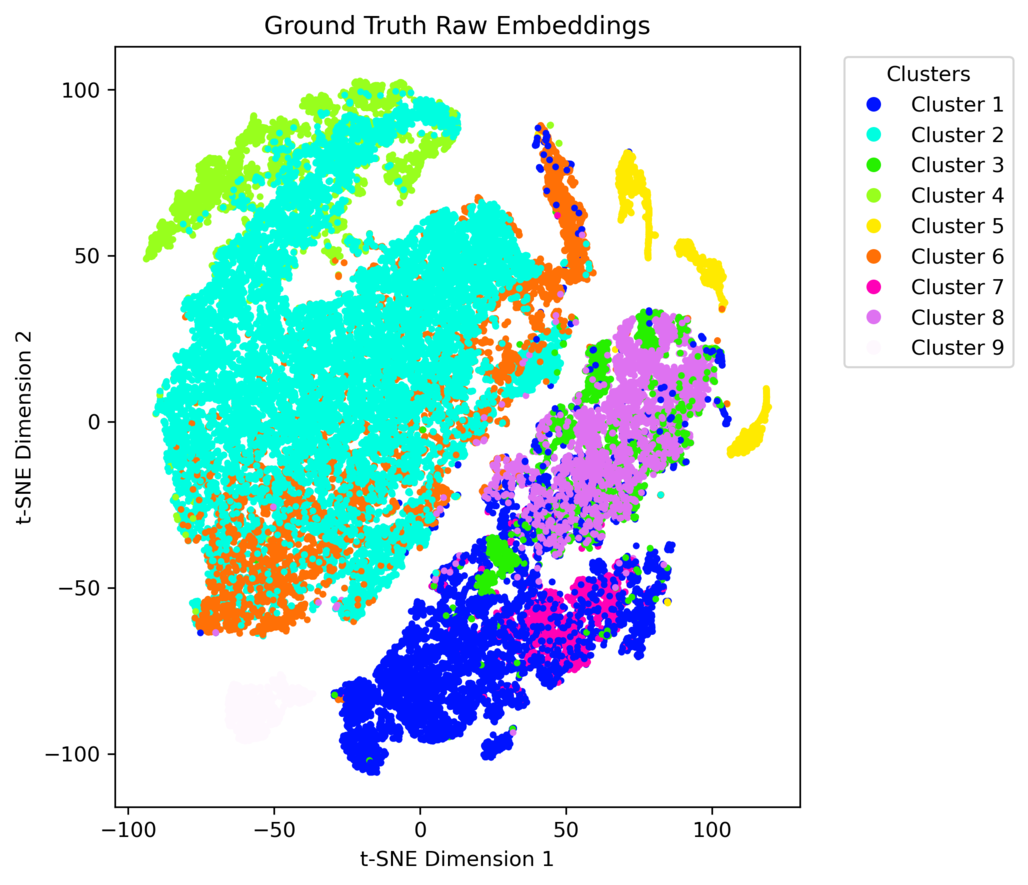}}
&
\raisebox{0.25\height}{\includegraphics[width=0.12\textwidth]{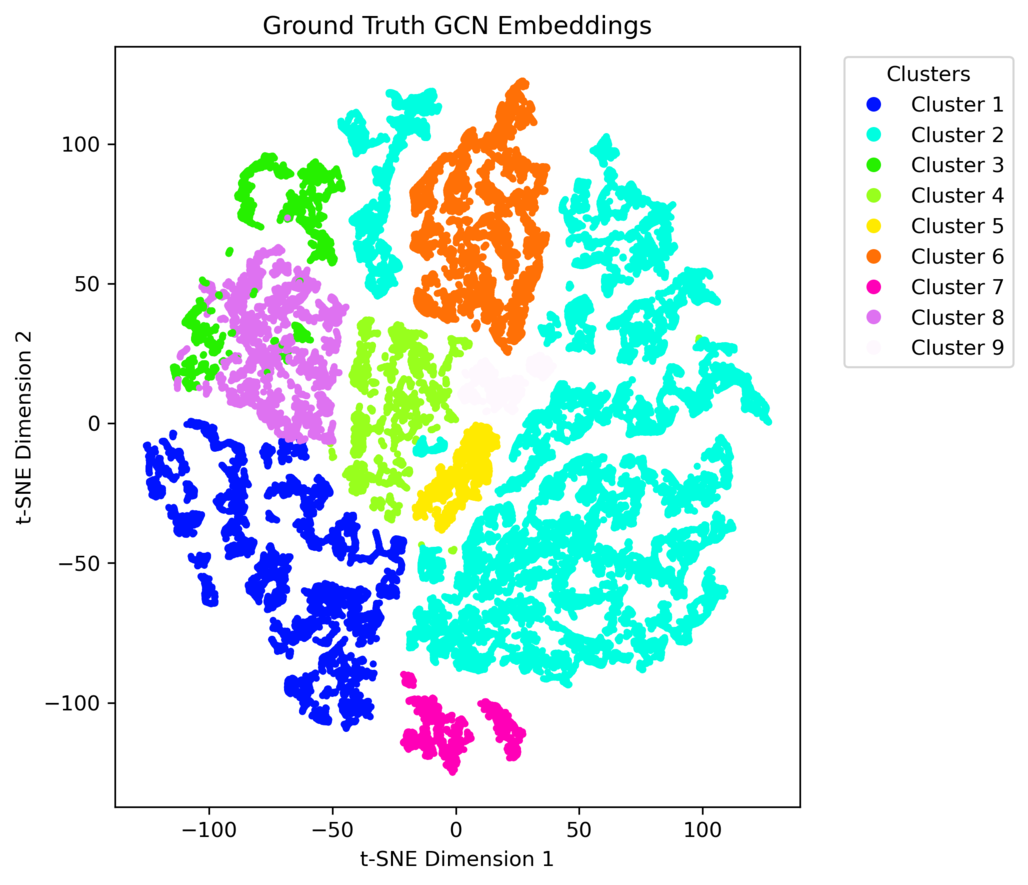}}
&
\raisebox{0.25\height}{\includegraphics[width=0.12\textwidth]{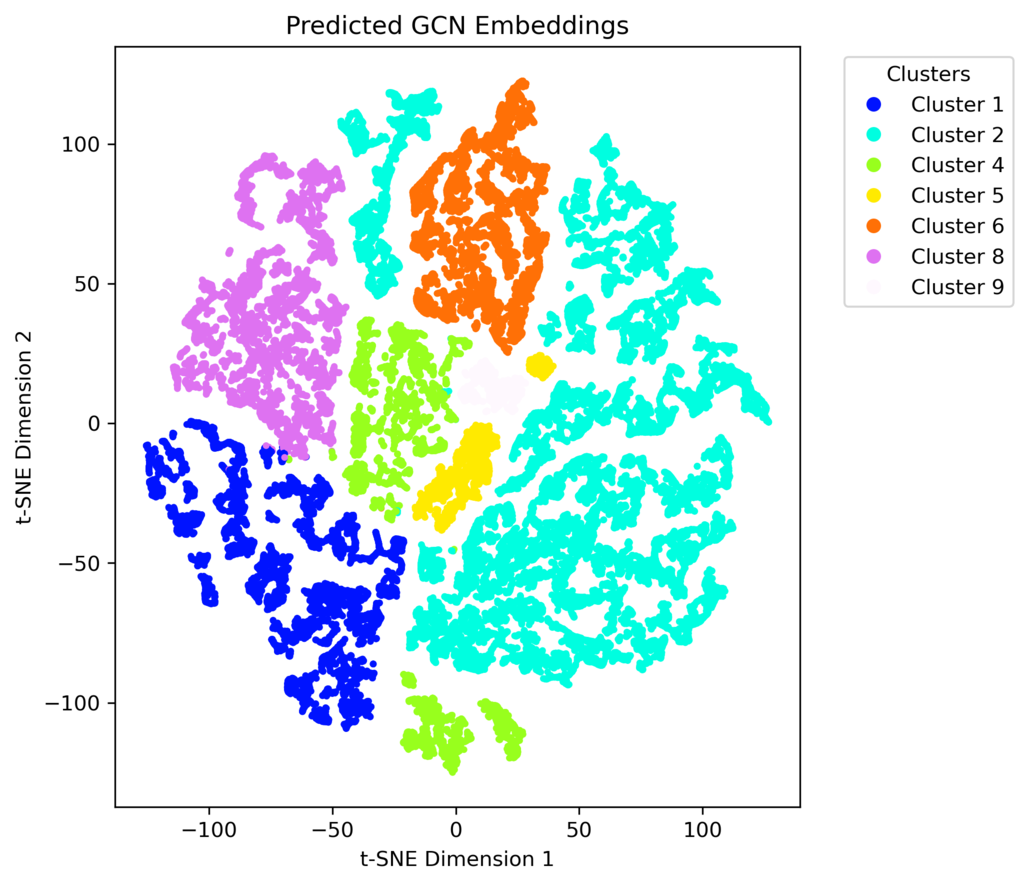}}
&
\includegraphics[width=0.12\textwidth]{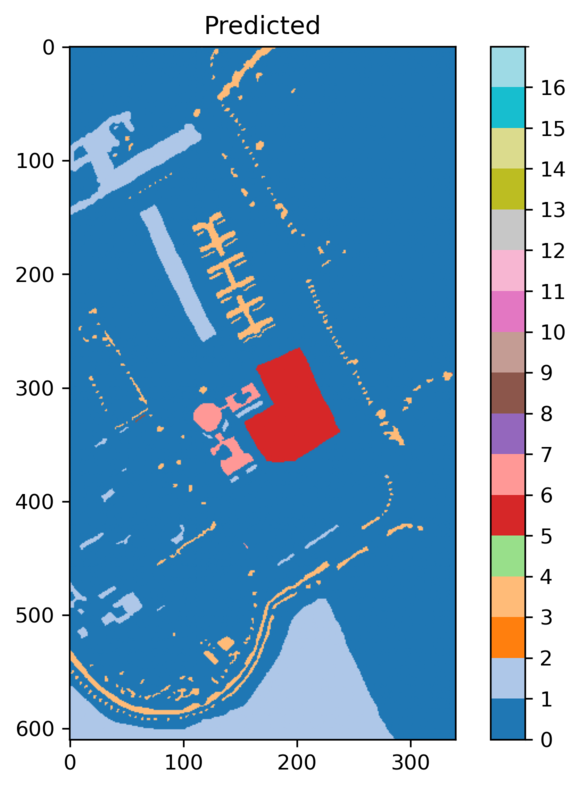}
\\ 
& \centering \textbf{DB:} 1.9341 & \centering \textbf{DB:} 1.7647 & \centering \textbf{DB:} 1.8759 & \\
& \centering \textbf{SH:} 0.0774 & \centering \textbf{SH:} 0.1416 & \centering \textbf{SH:} 0.169 & \\
\hline
\raisebox{3\height}{
\rotatebox[origin=c]{90}{\textbf{Salinas}}
\hspace{1mm}}
\includegraphics[width=0.12\textwidth]{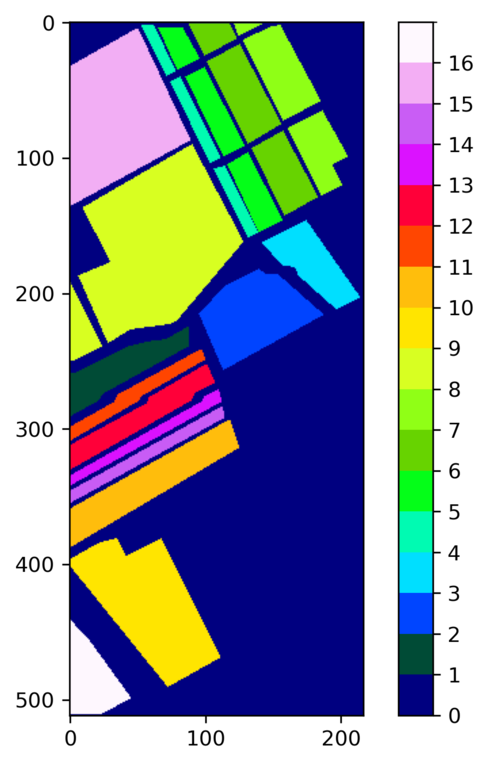}
&
\raisebox{0.35\height}{\includegraphics[width=0.12\textwidth]{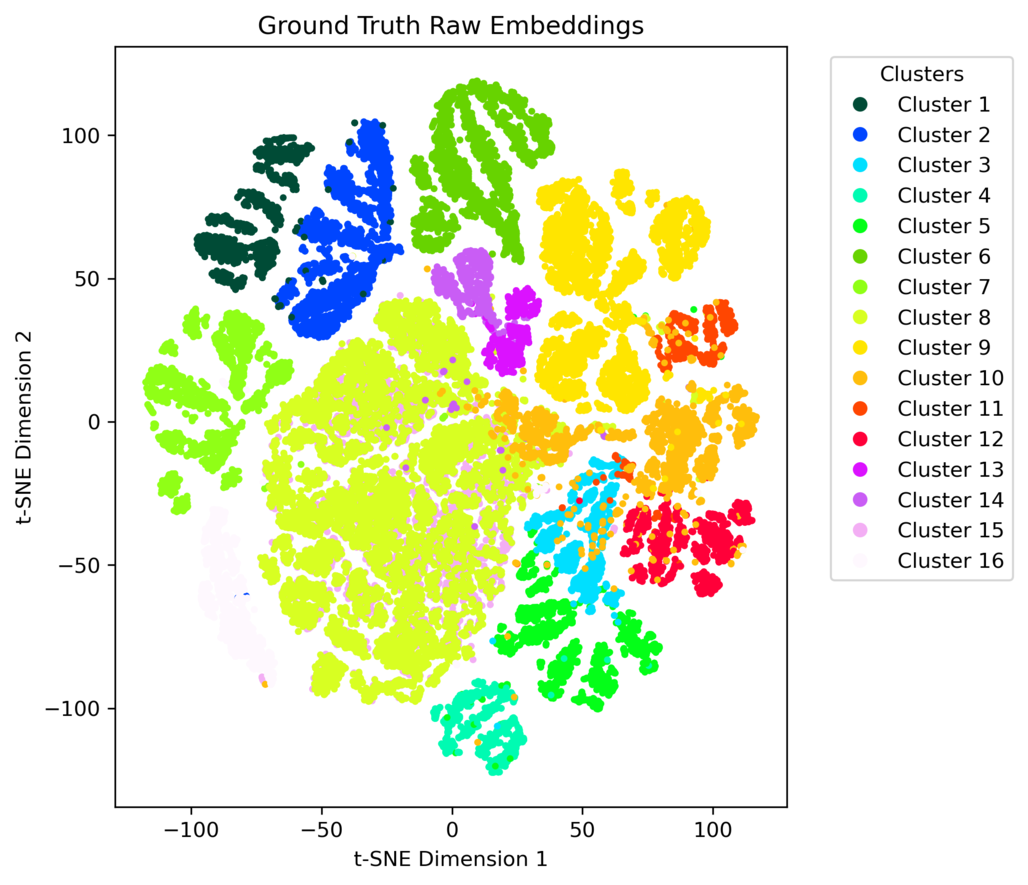}}
&
\raisebox{0.35\height}{\includegraphics[width=0.12\textwidth]{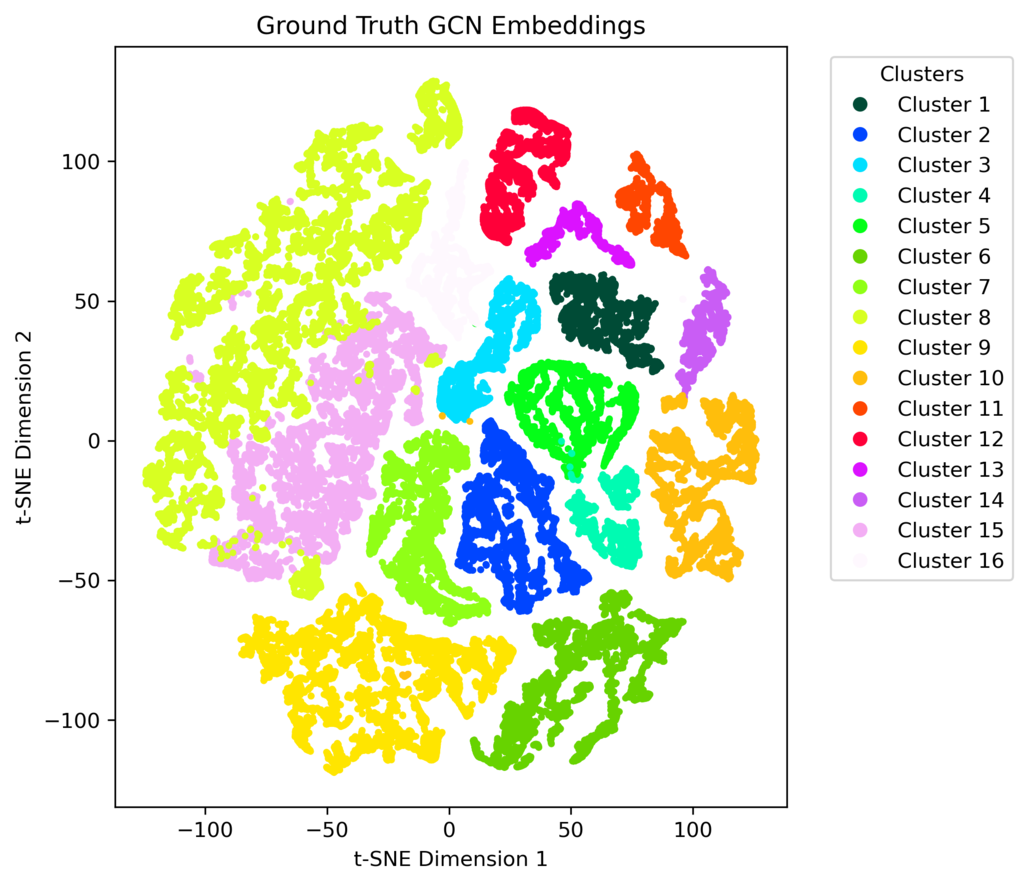}}
&
\raisebox{0.35\height}{\includegraphics[width=0.12\textwidth]{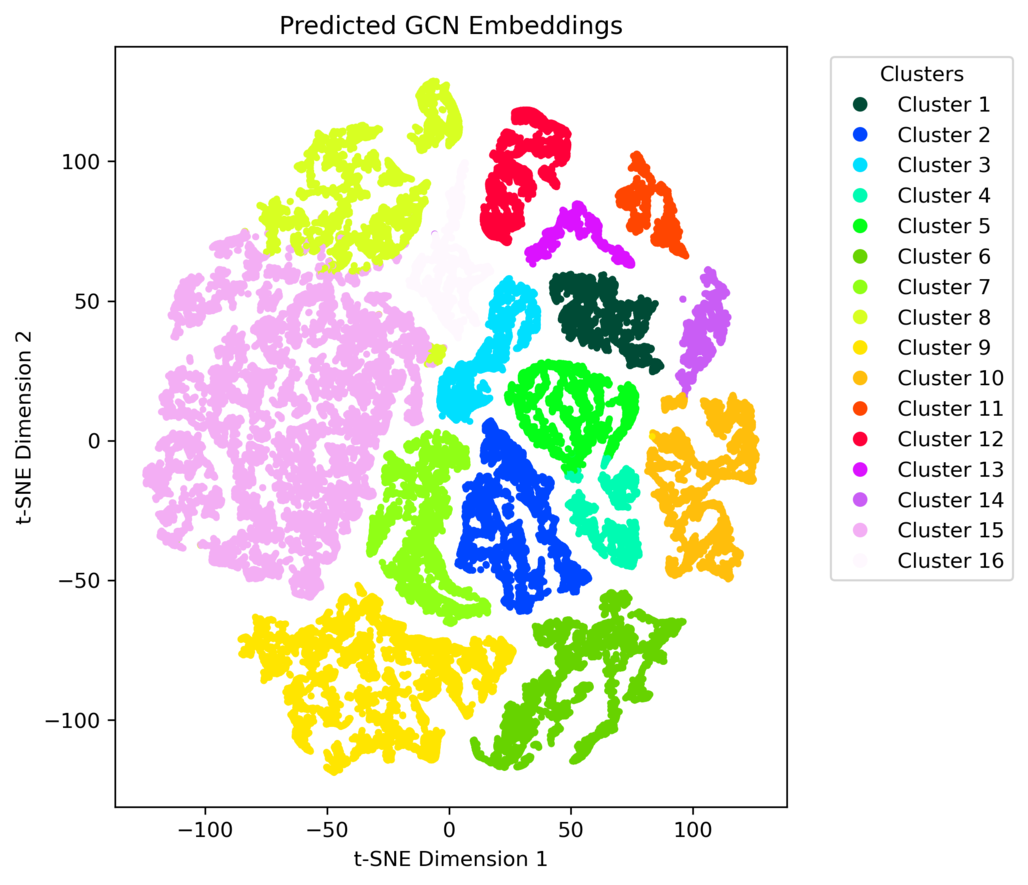}}
&
\includegraphics[width=0.12\textwidth]{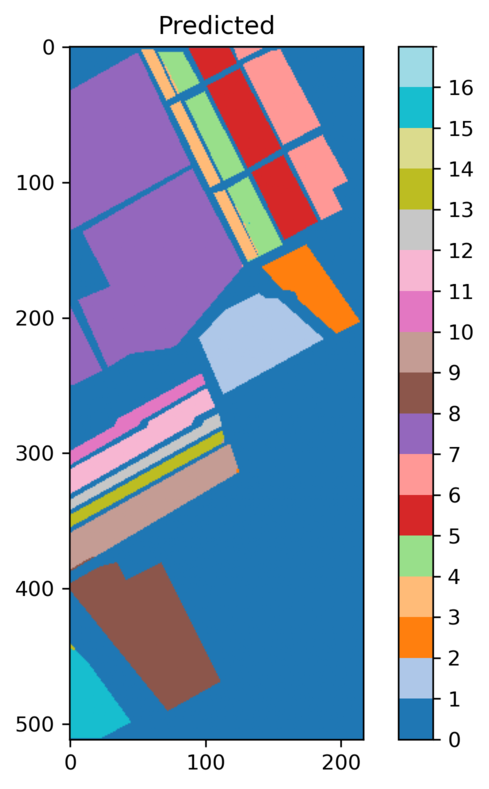}
\\ 
& \centering \textbf{DB:} 1.9338 & \centering \textbf{DB:} 1.2215 & \centering \textbf{DB:} 1.0451 & \\
& \centering \textbf{SH:} 0.237 & \centering \textbf{SH:} 0.3953 & \centering \textbf{SH:} 0.4112 & \\
\hline
\raisebox{2.5\height}{
\rotatebox[origin=c]{90}{\textbf{HongHu}}
\hspace{1mm}}
\includegraphics[width=0.12\textwidth]{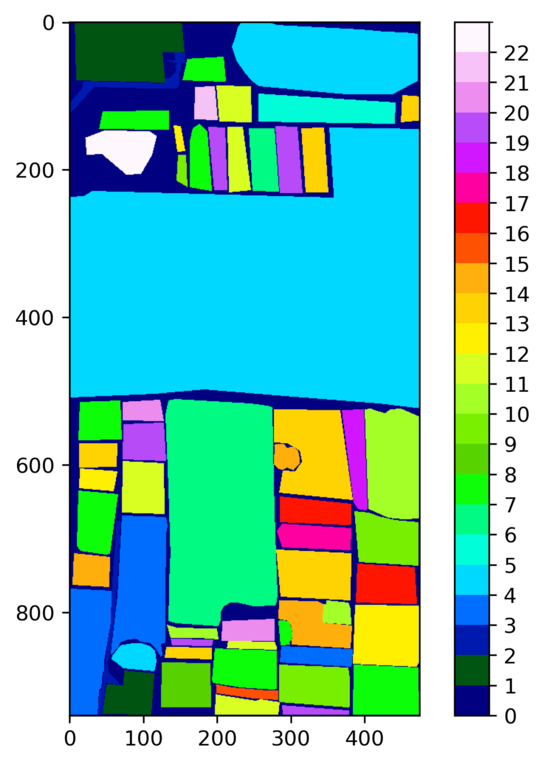}
&
\raisebox{0.3\height}{\includegraphics[width=0.12\textwidth]{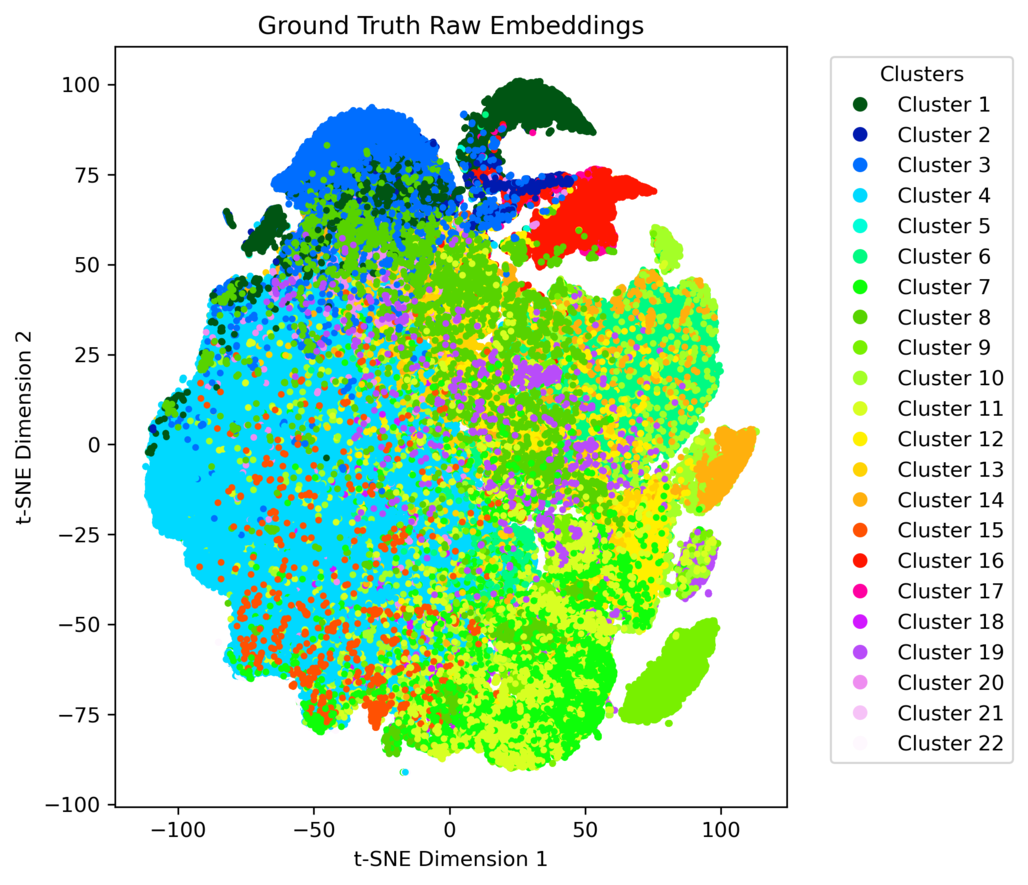}}
&
\raisebox{0.3\height}{\includegraphics[width=0.12\textwidth]{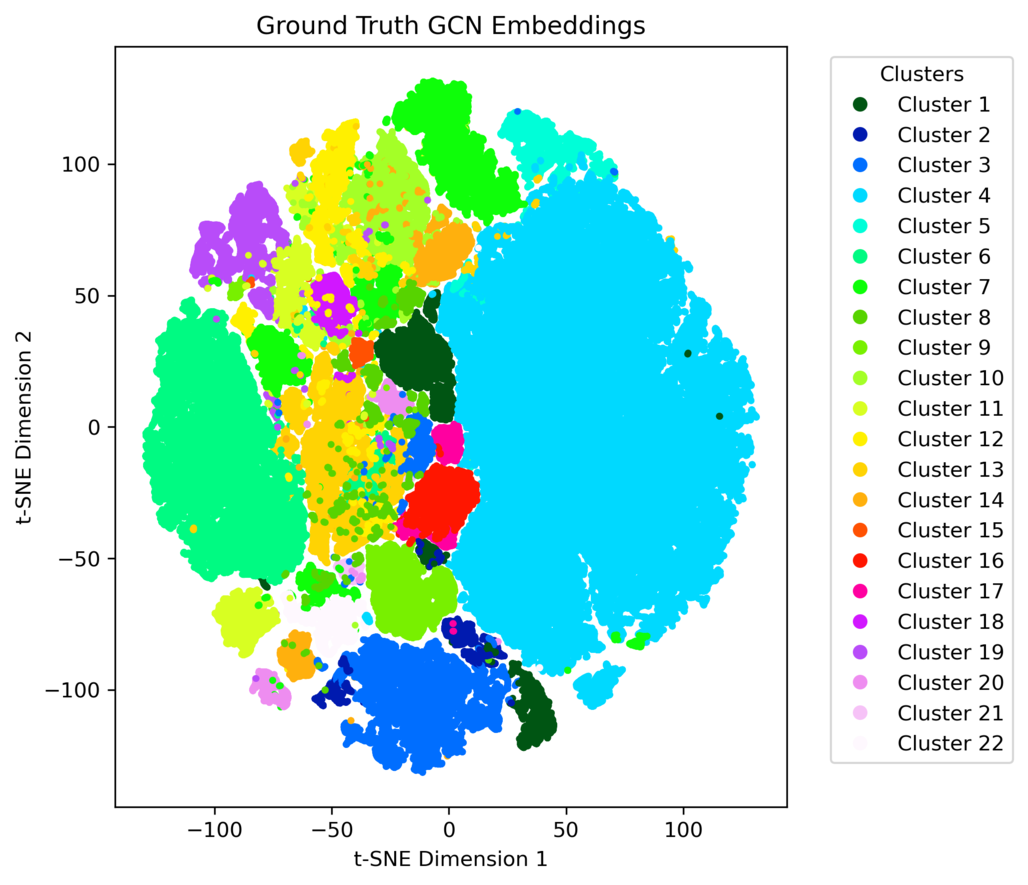}}
&
\raisebox{0.3\height}{\includegraphics[width=0.12\textwidth]{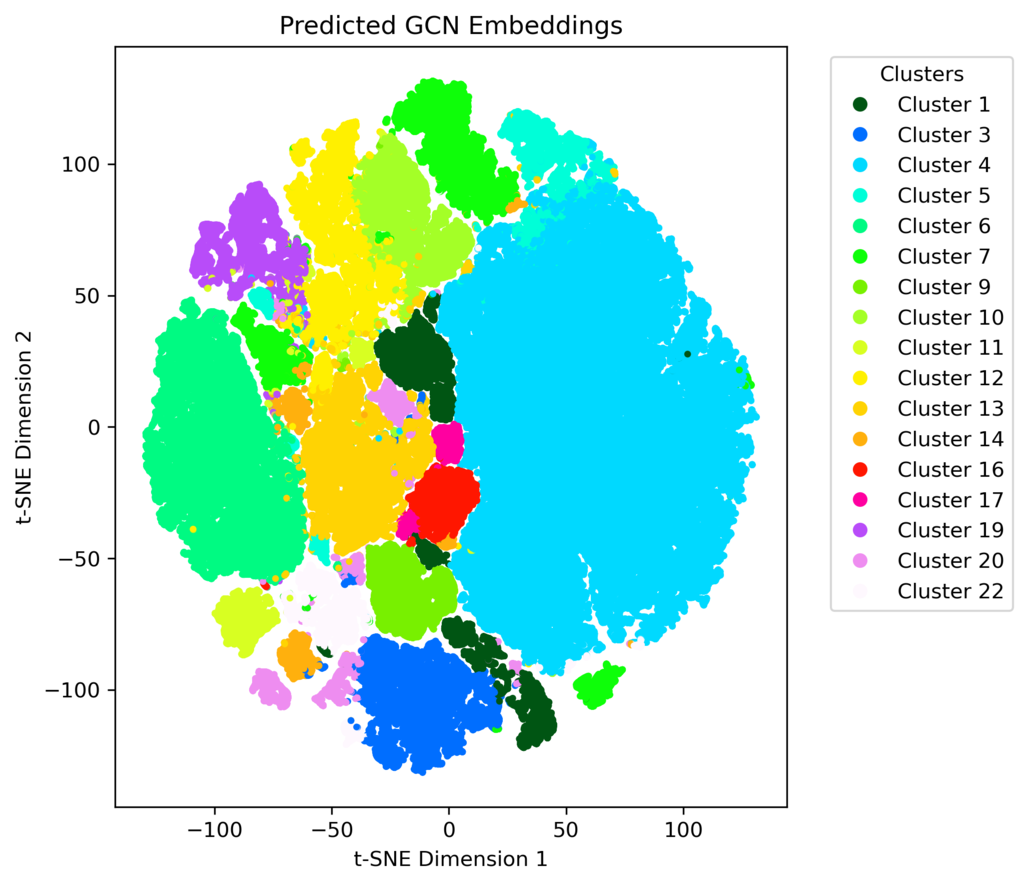}}
&
\includegraphics[width=0.12\textwidth]{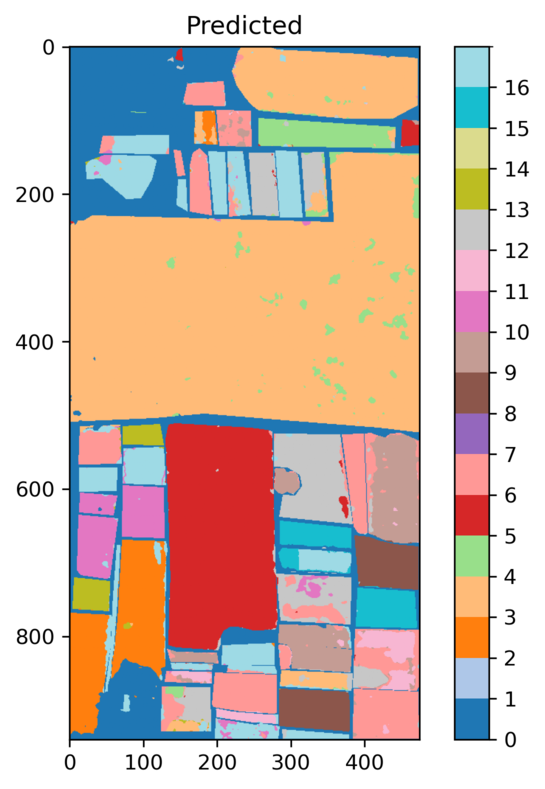}
\\ 
& \centering \textbf{DB:} 7.5796 & \centering \textbf{DB:} 2.2037 & \centering \textbf{DB:} 1.865 & \\
& \centering \textbf{SH:} -0.0436 & \centering \textbf{SH:} 0.0806 & \centering \textbf{SH:} 0.1494 & \\
\hline
\raisebox{3.8\height}{
\rotatebox[origin=c]{90}{\textbf{Botswana}}
\hspace{1mm}}
\includegraphics[width=0.12\textwidth]{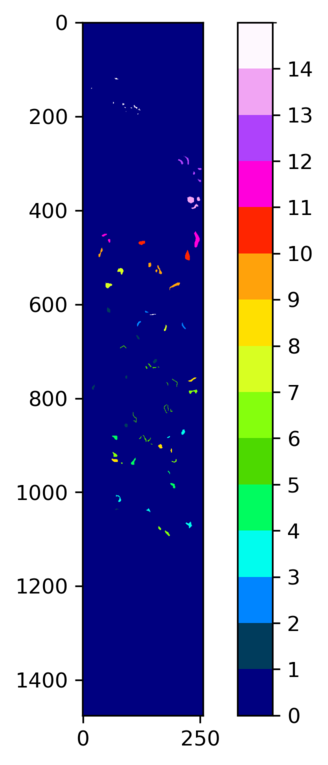}
&
\raisebox{0.8\height}{\includegraphics[width=0.12\textwidth]{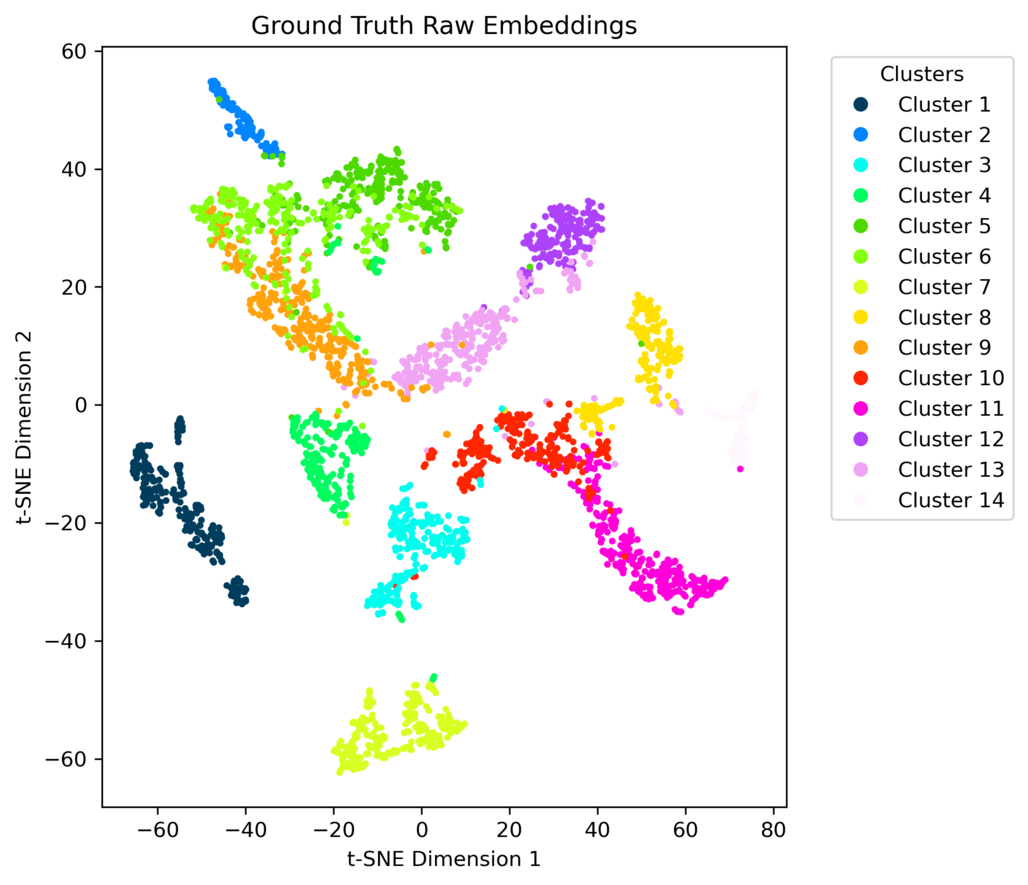}}
&
\raisebox{0.8\height}{\includegraphics[width=0.12\textwidth]{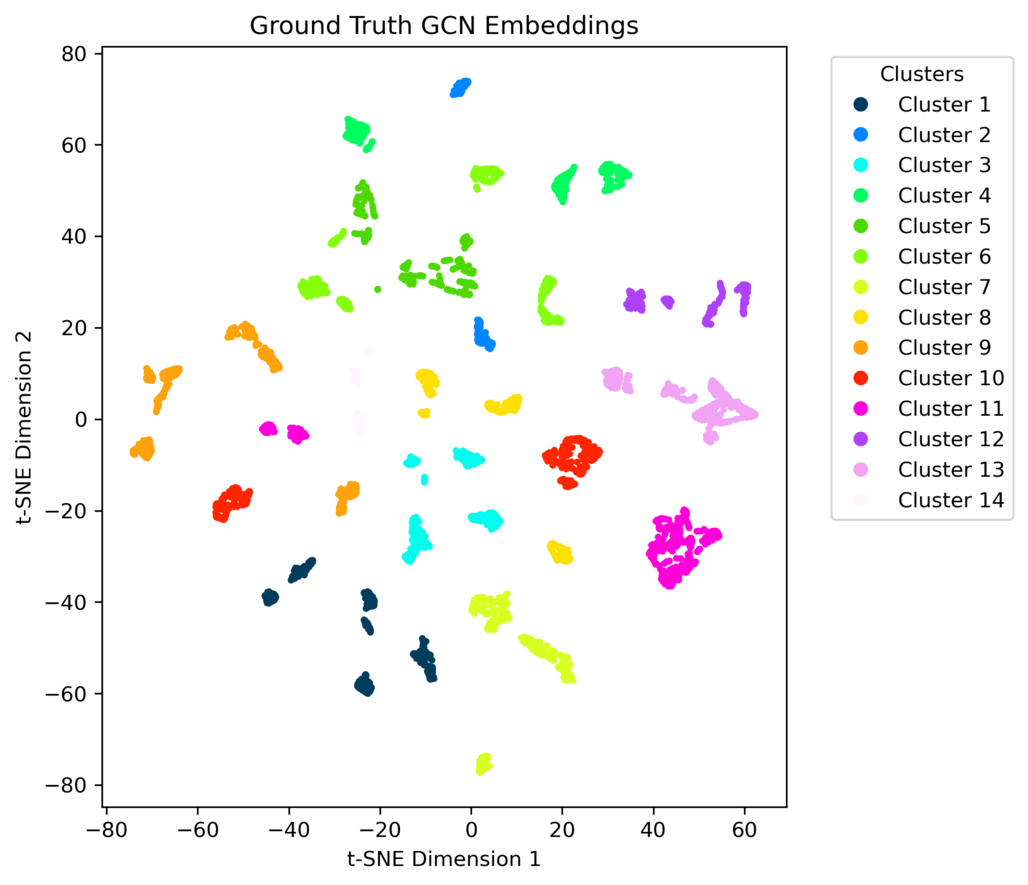}}
&
\raisebox{0.8\height}{\includegraphics[width=0.12\textwidth]{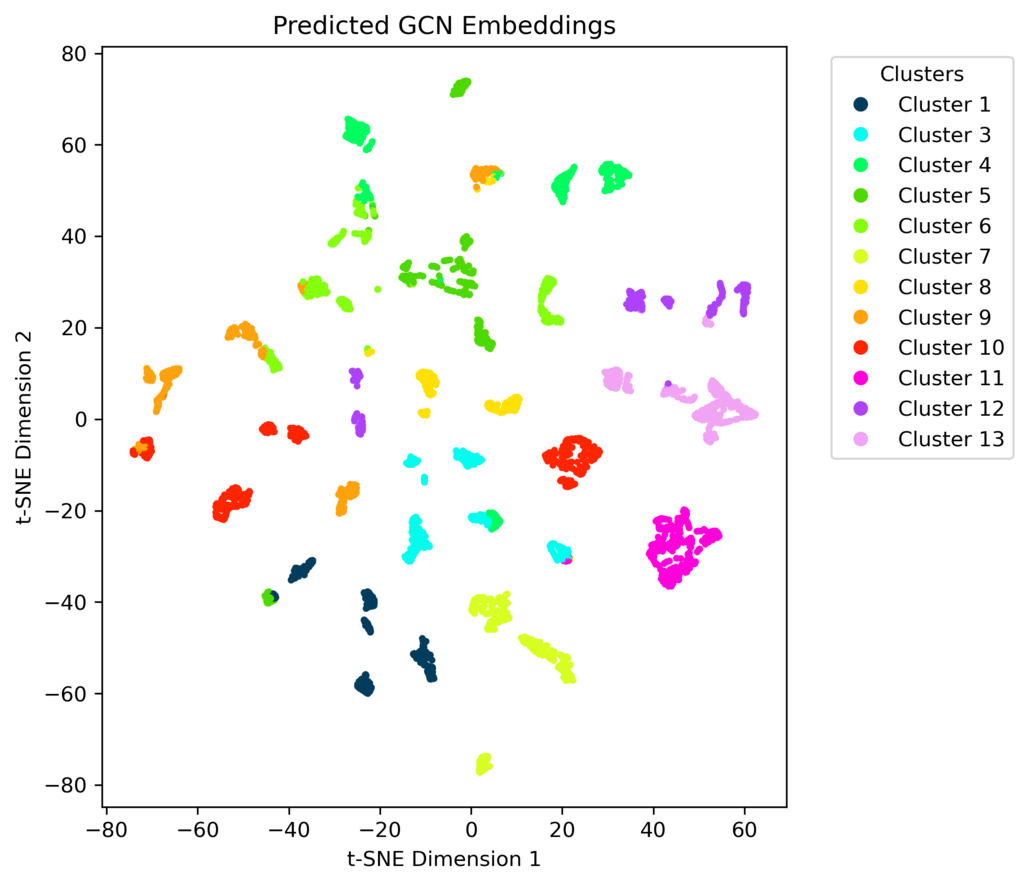}}
&
\includegraphics[width=0.12\textwidth]{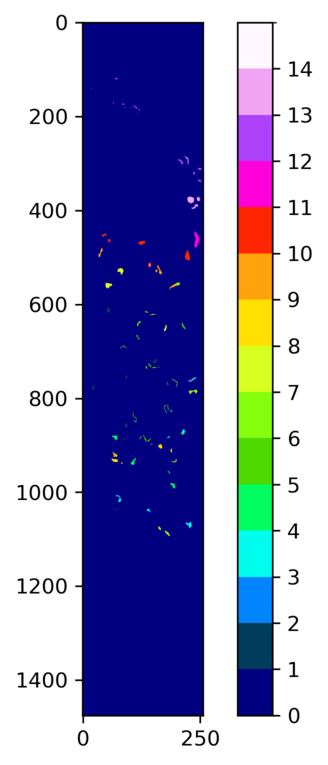}
\\ 
& \centering \textbf{DB:} 1.3903 & \centering \textbf{DB:} 1.3906 & \centering \textbf{DB:} 1.469 & \\
& \centering \textbf{SH:} 0.209 & \centering \textbf{SH:} 0.3339 & \centering \textbf{SH:} 0.3249 & \\
\hline
\raisebox{2.5\height}{
\rotatebox[origin=c]{90}{\textbf{KSC}}
\hspace{1mm}}
\includegraphics[width=0.12\textwidth]{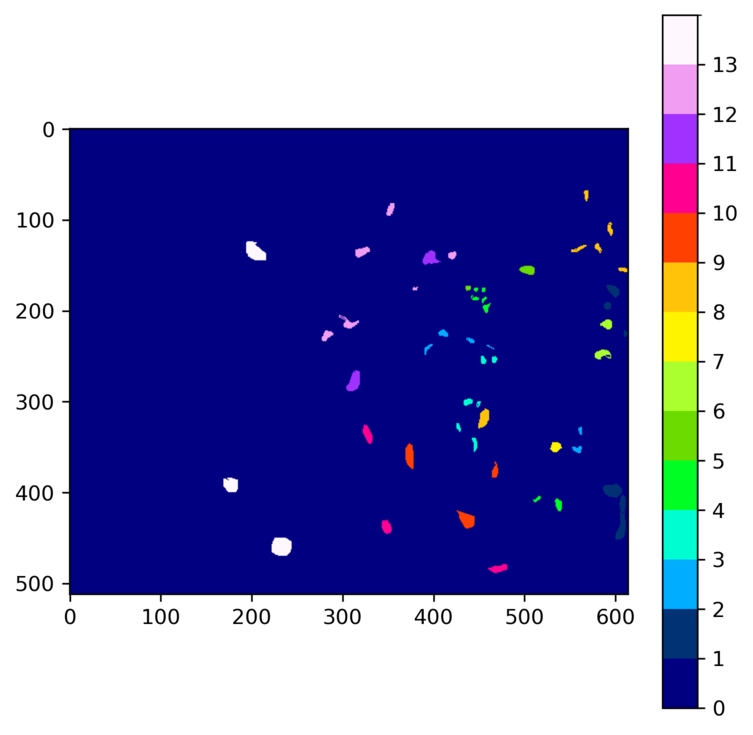}
&
\raisebox{0.05\height}{\includegraphics[width=0.12\textwidth]{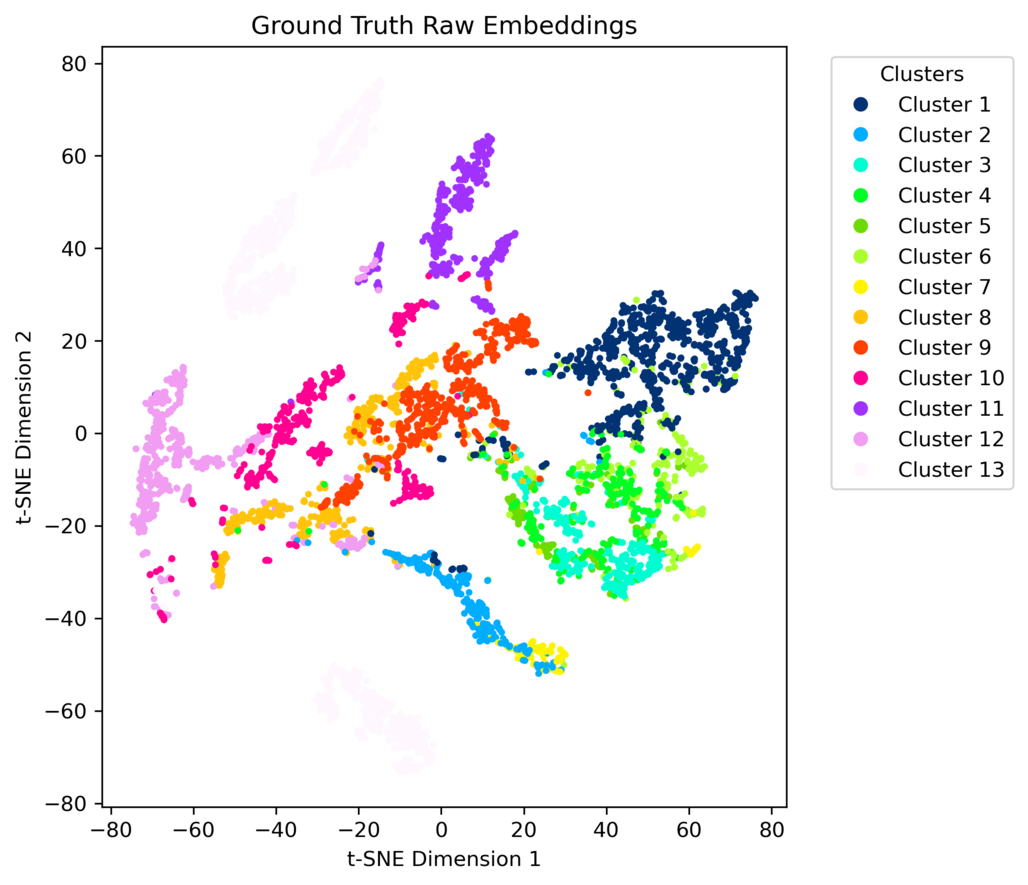}}
&
\raisebox{0.05\height}{\includegraphics[width=0.12\textwidth]{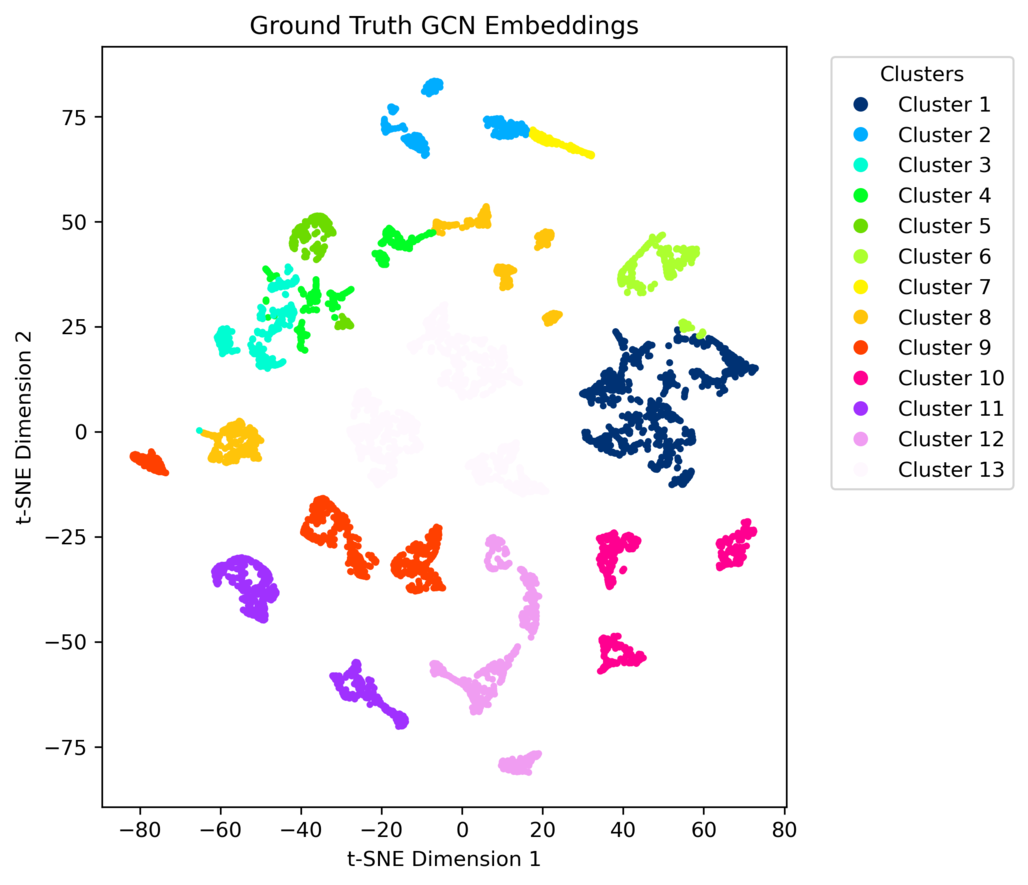}}
&
\raisebox{0.05\height}{\includegraphics[width=0.12\textwidth]{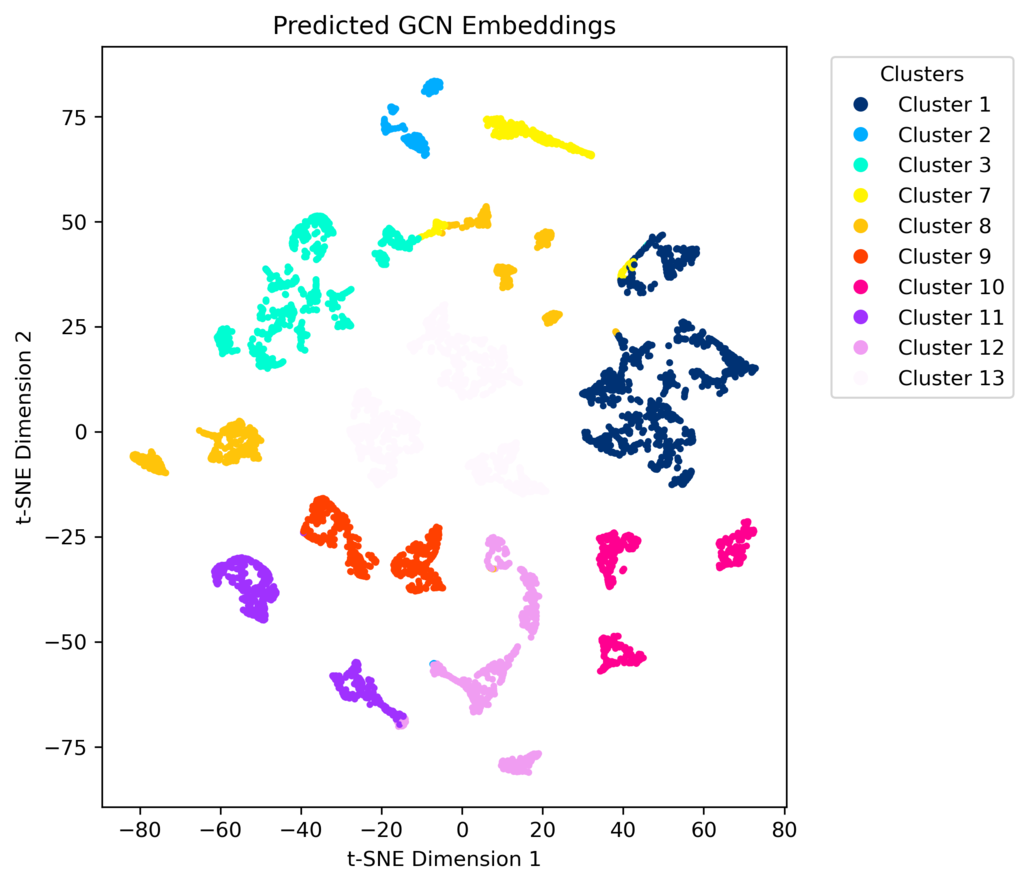}}
&
\includegraphics[width=0.12\textwidth]{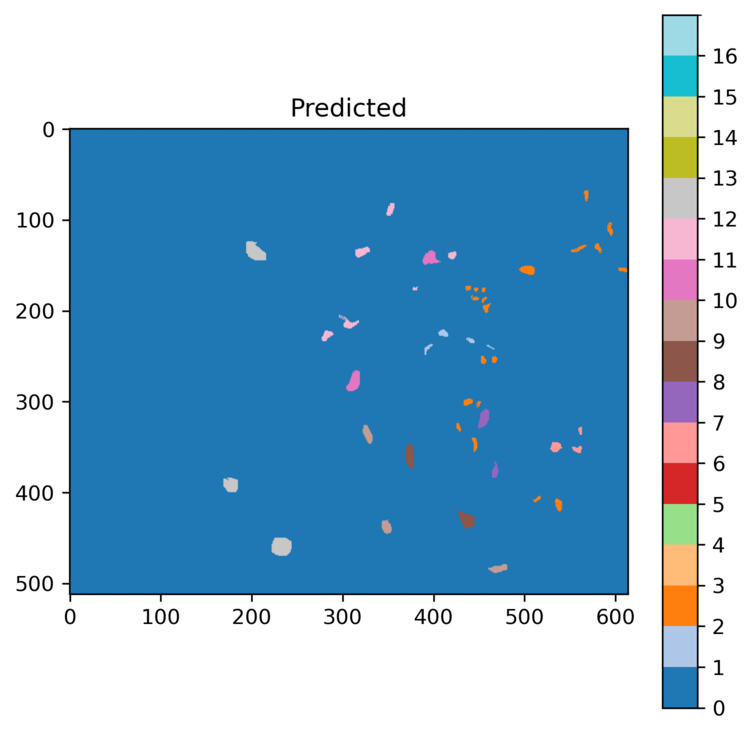}
\\ 
& \centering \textbf{DB:} 2.7933 & \centering \textbf{DB:} 1.6626 & \centering \textbf{DB:} 1.2038 & \\
& \centering \textbf{SH:} 0.1914 & \centering \textbf{SH:} 0.3448 & \centering \textbf{SH:} 0.4288 & \\
\hline
\end{tabular}
}
\end{table}
\begin{table*}[ht]
\centering
\caption{Quantitative performance of the proposed framework across all datasets.}
\label{tab:QCR}
\resizebox{\textwidth}{!}{%
\begin{tabular}{l|p{0.5cm}p{0.6cm}|p{0.6cm}p{0.6cm}p{0.6cm}p{0.7cm}|p{1.3cm}p{1.4cm}p{0.6cm}p{0.7cm}|p{0.6cm}p{0.6cm}p{1.1cm}p{0.7cm}|p{1.6cm}p{1.9cm}p{1.4cm}p{1.1cm}}
\hline 
\textbf{Dataset} & \multicolumn{10}{c|}{\textbf{External Index}} & \multicolumn{4}{c|}{\textbf{Internal Index}} & \multicolumn{4}{c}{\textbf{Efficiency}} \\\cline{2-19} 
& \textbf{OA} & \textbf{AA} & \textbf{NMI} & \textbf{ARI} & \textbf{AMI} & \textbf{FMI} & \textbf{Homogeneity} & \textbf{Completeness} & \textbf{Kappa} & \textbf{Purity} & \textbf{DB} & \textbf{SH} & \textbf{CH} & \textbf{Dunn} & \textbf{Total Inference Time (s)} & \textbf{Inference Time / Sample (s)} & \textbf{Peak Memory (MB)} & \textbf{FLOPS / Sample} \\ \hline 
\textbf{IndianP} &  80.49 & 57.41 & 0.7818 & 0.7143 & 0.7811 & 0.7534 & 0.7559 & 0.8095 & 0.7767 & 0.8049 & 1.8208 & 0.206 & 736.7607 & 0.0358 & 5.1013 & 2.4263e-4 & 29.54 & 5.13e+07 \\
\textbf{PaviaU} &  90.61 & 74.76 & 0.9206 & 0.9446 & 0.9205 & 0.9587 & 0.8815 & 0.9633 & 0.8753 & 0.9061 & 1.8759 & 0.169 & 3371.7265 & 0.0515 & 6.0811 & 2.932e-5 & 81.59 & 3.77e+08 \\
\textbf{Salinas} &  86.69 & 95.74 & 0.9279 & 0.7124 & 0.9279 & 0.745 & 0.9201 & 0.9359 & 0.8533 & 0.8669 & 1.0451 & 0.4112 & 7823.4914 & 0.0182 & 22.3054 & 2.0076e-4 & 155.14 & 1.43e+09 \\
\textbf{HongHu} &  84.46 & 62.11 & 0.7964 & 0.9066 & 0.7963 & 0.9259 & 0.7891 & 0.8038 & 0.8049 & 0.8446 & 1.865 & 0.1494 & 15159.9469 & - & 7.3074 & 2.1793e-4 & 1409.52 & 1.30e+11 \\
\textbf{Botswana} &  80.7 & 74.02 & 0.8214 & 0.7039 & 0.8198 & 0.7287 & 0.8022 & 0.8414 & 0.7904 & 0.807 & 1.469 & 0.3249 & 492.2974 & 0.0323 & 2.6438 & 7.00e-6 & 9.75 & 1.33e+05 \\
\textbf{KSC} &  83.13 & 71.76 & 0.885 & 0.8316 & 0.8844 & 0.854 & 0.8443 & 0.9298 & 0.8115 & 0.8313 & 1.2038 & 0.4288 & 1175.4554 & 0.0247 & 1.6551 & 5.26e-6 & 14.36 & 4.94e+05 \\
 \hline
\end{tabular}%
}
\end{table*}
Table~\ref{tab:visual_results} and Table \ref{tab:QCR} show the qualitative and quantitative performance, respectively, of the proposed framework and are analyzed simultaneously in this section. OA and AA provide insight into category-wise balance in performance, and for IndianP and HongHu, the gap is $23$\% and $22$\%, respectively, due to LULC imbalance. The smaller regions belonging to clusters $1-4$, $10-12$, and $15-16$ of IndianP became part of spectrally similar neighboring bigger regions in the latent space, which reduces AA. For HongHu, the embeddings of different categories are not sufficiently well separated in the latent space, which exacerbates the situation and makes smaller clusters (e.g., clusters $8$ and $14$) merge with larger clusters. With partial bias toward dominant categories and misidentification of minority clusters $4$ and $7$, for PaviaU, the gap is $16$\%. Due to lower LULC imbalance, Botswana and KSC have a relatively small gap ($6$\%). The situation reverses for Salinas, where smaller cluster regions are preserved well, but larger clusters ($8$ and $15$) have high intra-cluster variability and lower separability.

Proposed work achieves identical NMI and AMI values across all datasets, indicating statistically reliable clustering performance with minimal chance agreement. Whereas significantly high FMI and ARI values signify that pixels from the same cluster remain together and those from different clusters are separated with respect to GT. From Table~\ref{tab:QCR} it can be observed that for PaviaU and KSC, NMI and ARI are almost similar, indicating balanced semantic preservation and pairwise consistency, which is also evident from the comparative visualizations in $3$-rd and $4$-th columns of Table \ref{tab:visual_results} for the respective datasets. For IndianP and Botswana, the separation between NMI and ARI comparatively increases, as the visualizations show some clusters (i.e., $5$, $10$, and $12$ both in Botswana and IndianP GT) have merged with nearby clusters in predicted segmentations. Due to boundary ambiguity between clusters $8$ and $15$ the NMI and ARI differ significantly for Salinas. For HongHu, although semantic information and category distributions are preserved, the boundary separation is much less in latent space, so the NMI is lower than ARI.

For all datasets, homogeneity is lower than completeness, which indicates same-class samples stay together, but clusters may contain mixed classes. This behavior is desirable in HSI segmentation, where maintaining semantic continuity is often preferred over highly fragmented yet pure clusters. The kappa score for PaviaU and Salinas is higher than the others due to higher homogeneity and completeness. For other datasets, either one of them or both drop and reduce the Kappa score. Purity generally follows the trend of homogeneity since both reward intra-cluster consistency, but if smaller clusters (like the cluster $8$ in Salinas) are included in a bigger cluster, then purity drops. Homogeneity is higher than completeness in KSC as clusters $7$ and $9$, $3$ and $10$, $9$ and $10$ collapse.

Internal clustering metrics provide insight into the structure of the learned latent space. Lower DB values and higher SH scores for Salinas and KSC indicate compact, well-separated clusters, reflecting a well-organized embedding space. Although Salinas exhibits the strongest overall cluster organization, its relatively low Dunn index suggests localized overlap between a small number of cluster pairs (e.g., clusters $8$ and $14$). IndianP is characterized by high DB, low SH, and moderate CH, indicating moderate inter-class overlap despite the formation of distinct clusters. For PaviaU, low DB and SH combined with high CH and Dunn suggest strong global cluster separation, although some local overlap persists, consistent with the external evaluation metrics. In contrast, HongHu exhibits high DB, low SH, and extremely high CH, indicating large between-cluster variance but poor local separation, with substantial overlap near cluster boundaries. Botswana demonstrates balanced clustering characteristics, with moderate DB, comparatively higher SH than IndianP and PaviaU, and lower CH, suggesting reasonably compact clusters with moderate global separation. Finally, all datasets except HongHu (for which the Dunn index could not be computed because of the computational cost of pairwise distance calculations) exhibit relatively low Dunn values, indicating that while the learned embedding space preserves overall semantic structure, localized overlap remains at a few class boundaries.
\begin{table*}[hb]
\centering
\caption{
Performance comparison with $10$ SOTA methods on $4$ benchmark HSI datasets with best and second best performance represented in \textcolor{blue}{\textbf{blue}} and \textcolor{teal}{\textbf{teal}}. One-sided t-test ($p < 0.05$) is done for statistical significance.}
\label{table_3}
\resizebox{\textwidth}{!}{
\begin{tabular}{c|c|cccccccccc|ccc}
\hline
\textbf{Datasets} & \textbf{Metrics} & \textbf{FSCAG \cite{wang2017fast}} & \textbf{SGCNR \cite{wang2019scalable}} & \textbf{HESSC \cite{rafiezadeh2020hierarchical}} & \textbf{NCSC \cite{cai2022superpixel}} & \textbf{SGLSC \cite{zhao2021superpixel}} & \textbf{S$^3$AGC \cite{chen2023spectral}} & \textbf{DSCRLE \cite{zhao2023deep}} & \textbf{BGPC \cite{zhang2024bipartite}} & \textbf{SAPC \cite{jiang2024structured}} & \textbf{EGFSC \cite{zhang2025elastic}} & \textbf{Proposed $\pm$ SD} & \textbf{t-statistic}	& \textbf{p-value} \\ \hline
\multirow{5}{*}{IndianP} 
 & OA & 36.13 & 35.75 & 34.08 & 58.65 & 59.50 & 49.94 & 47.80 & 43.85 & 55.88 & \textcolor{teal}{\textbf{70.08}} & \textcolor{blue}{\textbf{80.49 $\pm$ 0.0728}} & 315.75 & 3.02e-10 \\
 & AA & 37.82 & 35.35 & 35.12 & 49.46 & 54.25 & 42.25 & 39.88 & 41.84 & 52.39 & \textcolor{teal}{\textbf{55.74}} & \textcolor{blue}{\textbf{57.41 $\pm$ 0.0683}} & 50.65  & 4.55e-07 \\
 & Kappa & 0.2955 & 0.2889 & 0.2866 & 0.5346 & 0.5522 & 0.4317 & 0.3996 & 0.3782 & 0.4959 & \textcolor{teal}{\textbf{0.6580}} & \textcolor{blue}{\textbf{0.7767 $\pm$ 0.0009}} & 303.93 & 3.52e-10 \\
 & NMI & 0.4513 & 0.4463 & 0.4133 & 0.5911 & 0.6190 & 0.5157 & 0.4779 & 0.5020 & 0.5403 & \textcolor{teal}{\textbf{0.6910}} & \textcolor{blue}{\textbf{0.7818 $\pm$ 0.0019}} & 105.28 & 2.44e-08 \\
 & Purity & 0.5187 & 0.5261 & 0.5042 & 0.6404 & 0.7058 & 0.5548 & 0.5270 & 0.5530 & 0.5933 & \textcolor{teal}{\textbf{0.7517}} & \textcolor{blue}{\textbf{0.8049 $\pm$ 0.0007}} & 159.41 & 4.64e-09 \\ \hline
 
\multirow{5}{*}{PaviaU} 
 & OA & 49.34 & 57.45 & 45.03 & 54.88 & 62.45 & 53.38 & 68.46 & 74.05 & 63.94 & \textcolor{teal}{\textbf{75.76}} & \textcolor{blue}{\textbf{90.61 $\pm$ 1.5697}} & 17.15 & 3.39e-05 \\
 & AA & 56.41 & 56.91 & 40.14 & 52.46 & 44.17 & 56.61 & \textcolor{teal}{\textbf{63.43}} & 61.88 & 59.20 & 55.41 & \textcolor{blue}{\textbf{74.76 $\pm$ 0.1252}} & 206.32 & 1.66e-09 \\
 & Kappa & 0.3937 & 0.4728 & 0.3468 & 0.4578 & 0.4885 & 0.4206 & 0.5907 & 0.6405 & 0.5294 & \textcolor{teal}{\textbf{0.6807}} & \textcolor{blue}{\textbf{0.8753 $\pm$ 0.0235}} & 14.50 & 6.58e-05 \\
 & NMI & 0.4962 & 0.5393 & 0.4799 & 0.5135 & 0.5410 & 0.4912 & 0.6351 & \textcolor{teal}{\textbf{0.6900}} & 0.6189 & 0.6679 & \textcolor{blue}{\textbf{0.9206 $\pm$ 0.0148}} & 30.97 & 3.24e-06 \\
 & Purity & 0.6601 & 0.6981 & 0.6268 & 0.6775 & 0.6769 & 0.6404 & 0.7241 & 0.7406 & 0.7132 & \textcolor{teal}{\textbf{0.7857}} & \textcolor{blue}{\textbf{0.9061 $\pm$ 0.0157}} & 13.15 & 9.65e-05 \\ \hline
 
\multirow{5}{*}{Salinas} 
 & OA & 68.14 & 68.81 & 52.10 & 78.26 & 76.64 & 74.65 & 72.79 & 78.80 & 78.81 & \textcolor{blue}{\textbf{87.28}} & \textcolor{teal}{\textbf{86.69 $\pm$ 0.15205}} & -12.68 & 0.9999 \\
 & AA & 61.92 & 66.36 & 47.47 & 60.93 & 65.47 & 71.54 & 71.55 & 72.69 & 73.73 & \textcolor{teal}{\textbf{76.61}} & \textcolor{blue}{\textbf{95.74 $\pm$ 1.0286}} & 37.59 & 1.50e-06 \\
 & Kappa & 0.6469 & 0.6491 & 0.4731 & 0.7566 & 0.7407 & 0.7181 & 0.6964 & 0.7632 & 0.7637 & \textcolor{blue}{\textbf{0.8577}} & \textcolor{teal}{\textbf{0.8533 $\pm$ 0.0031}} & -7.14 & 0.9990 \\
 & NMI & 0.7589 & 0.7813 & 0.6265 & 0.8546 & 0.8388 & 0.7836 & 0.8035 & 0.8134 & 0.8464 & \textcolor{teal}{\textbf{0.8918}} & \textcolor{blue}{\textbf{0.9279 $\pm$ 0.0086}} & 13.36 & 9.06e-05 \\
 & Purity & 0.7049 & 0.7054 & 0.5770 & 0.8012 & 0.7679 & 0.7521 & 0.7429 & 0.7931 & 0.8056 & \textcolor{blue}{\textbf{0.8740}} & \textcolor{teal}{\textbf{0.8669 $\pm$ 0.0015}} & -14.44 & 0.9999 \\ \hline

\multirow{5}{*}{HongHu} 
 & OA & 32.03 & 32.03 & 32.12 & 41.33 & 46.16 & 39.41 & 33.55 & 52.98 & 51.28 & \textcolor{teal}{\textbf{77.23}} & \textcolor{blue}{\textbf{84.46 $\pm$ 1.1953}} & 10.73 & 0.0002\\
 & AA & 35.88 & 30.34 & 25.27 & 45.06 & 51.09 & 28.66 & 40.41 & 31.56 & 40.26 & \textcolor{teal}{\textbf{53.52}} & \textcolor{blue}{\textbf{62.11 $\pm$ 5.1477}} & 0.45 & 0.3371\\
 & Kappa & 0.2700 & 0.2646 & 0.2573 & 0.3727 & 0.4143 & 0.3228 & 0.2935 & 0.4306 & 0.4364 & \textcolor{teal}{\textbf{0.7113}} & \textcolor{blue}{\textbf{0.8049 $\pm$ 0.01651}} & 9.71 & 0.0003\\
 & NMI & 0.3983 & 0.4275 & 0.3983 & 0.5855 & 0.6271 & 0.5091 & 0.5095 & 0.5243 & 0.5654 & \textcolor{teal}{\textbf{0.7282}} & \textcolor{blue}{\textbf{0.7964 $\pm$ 0.01006}} & 12.03 & 0.00014\\
 & Purity & 0.6369 & 0.6678 & 0.6369 & 0.7411 & 0.7856 & 0.7034 & 0.7210 & 0.7124 & 0.7452 & \textcolor{teal}{\textbf{0.8048}} & \textcolor{blue}{\textbf{0.8446 $\pm$ 0.0120}} & 4.65 & 0.0048\\ \hline
 \end{tabular}
}
\end{table*}

The 2D t-SNE visualizations for qualitative analysis of the proposed framework in Table~\ref{tab:visual_results} indicates that the latent representations learned by the MALNISI-GAT framework have a better separation than raw embeddings. It is true because DB reduces and SH improves in \textbf{H} with respect to \textbf{X} by learning difficult HSI manifolds. The clustering step further discriminates, except Botswana, which requires a dedicated sparse data handling component. The computational analysis demonstrates that the proposed framework maintains efficient inference behaviour across diverse HSI datasets with stable inference time per sample and dataset-dependent memory and computational requirements. Botswana and KSC achieve the lowest inference time, memory, and computational demand due to a small ROI with sparsity.

\textbf{B. Comparative Study with SOTA Methods: }

\textit{\textbf{1) Dense Datasets:}}
A comprehensive comparative analysis with SOTA methods over $5$ complementary metrics is presented in Table \ref{table_3} which shows an improvement over the best-performing method, EGFSC, of approximately $15$\%  and $12$\% in OA and NMI, respectively, on IndianP; $20$\%  and $38$\% in OA and NMI, respectively, on PaviaU; $4$\% improvement in NMI while maintaining highly competitive OA performance ($-0.04$\%) on Salinas; and $9$\% and $8$\% improvement in OA and NMI, respectively, on HonghHu. Over other metrics, the proposed framework achieves the best results on all datasets except purity ($-0.8$\%) and kappa ($-0.5$\%) in the Salinas dataset. The significant improvements in AA ($3$\%, $18$\%, $25$\%, and $17$\%) and kappa ($18$\%, $22$\%, $-0.5$\%, $13$\%), which assess the performance for every category without bias, show the discrimination capability. The high NMI and low standard deviation (SD) for all datasets indicate that the achieved segmentation has better agreement with GT categories and is stable. 
The one-sided t-test results reported in Table~\ref{table_3} yield $p$-values well below the $0.05$ threshold, confirming that the performance gains of the proposed framework over the strongest baselines are statistically significant across the dense datasets.

\textit{\textbf{2) Sparse Datasets:}}
There are methods that work dedicatedly for sparse datasets Botswana and Table \ref{tab:botswana_results} provides a comparative study with them. The proposed method achieves mostly second-best performance ($\sim 2$\% decrease) over all metrics, and the statistical significance shows the same. Now, it is true that the proposed method works competitively without any sparse data-addressing component, and on the other hand, these SOTA methods' performance drops more than ($\sim 10$\%) across all metrics when applied to dense datasets \cite{zhao2025deep}, and this shows the robustness and superiority of the proposed work.
\begin{table}[t]
\centering
\caption{Performance comparison on the Botswana dataset with $6$ SOTA methods, with best, second best, and third best performance represented in \textcolor{blue}{\textbf{blue}}, \textcolor{teal}{\textbf{teal}} and \textcolor{magenta}{\textbf{magenta}}. One-sided t-test ($p < 0.05$) is done for statistical significance.}
\resizebox{\linewidth}{!}{
\begin{tabular}{l|cccc}
\hline
\textbf{Method} & \textbf{OA} & \textbf{Kappa} & \textbf{NMI} & \textbf{Purity} \\
\hline
$K$-means \cite{zhou2022study}  & 64.56 & 0.6205 & 0.7416 & 0.6456 \\
H-GMM \cite{zhao2021robust}     & 66.41 & 0.6416 & 0.7493 & 0.6641 \\
FSEC \cite{9269402}             & 68.20 & 0.6579 & \textcolor{teal}{\textbf{0.8324}} & 0.6820 \\
EnSC \cite{you2016oracle}       & 65.64 & 0.6331 & 0.6886 & 0.6564 \\
SSC-BP \cite{elhamifar2013sparse} & \textcolor{magenta}{\textbf{70.91}} & \textcolor{magenta}{\textbf{0.6875}} & 0.7316 & \textcolor{magenta}{\textbf{0.7091}} \\
DSCFS \cite{zhao2025deep}       & \textcolor{blue}{\textbf{82.39}} & \textcolor{blue}{\textbf{0.8113}} & \textcolor{blue}{\textbf{0.8358}} & \textcolor{blue}{\textbf{0.8239}} \\
\hline
\textbf{Proposed $\pm$ SD} & \textcolor{teal}{\textbf{80.70 $\pm$ 0.1968}} & \textcolor{teal}{\textbf{0.7904 $\pm$ 0.0021}} & \textcolor{magenta}{\textbf{0.8214 $\pm$ 0.0012}} & \textcolor{teal}{\textbf{0.8070 $\pm$ 0.0020}} \\
\hline
\textbf{t-statistic} & $-20.20$ & $-23.72$ & $-27.67$ & $-20.20$ \\
\textbf{p-value}     & 0.99998  & 0.99999  & 0.99999  & 0.99998 \\
\hline
\end{tabular}
}
\label{tab:botswana_results}
\end{table}
KSC is another sparse dataset that has the latest developments in classification \cite{liu2025hyper}, band elimination \cite{zhang2025evaluate}, band extraction \cite{al2023fhic}, or band selection \cite{7112088}. The proposed work is one of the methods that performs segmentation with better performance (Tables \ref{tab:visual_results} and \ref{tab:QCR}) on this dataset.

\textbf{C. Impact of Other Similarity Indices on Segmentation: } 
This study evaluates the importance of the proposed MALNISI similarity measure by replacing it with alternative metrics, like Minkowski \cite{thompson1996minkowski}, SAM, SID, Dice \cite{li2020generic}, Mahalanobis distance \cite{gonzalez2009digital}, Structural Similarity Index Measure (SSIM) \cite{6467154}, and Scale Invariant Feature Transform (SIFT) \cite{6467154}, while preserving the remaining architecture. MALNISI demonstrates consistently strong and stable performance across IndianP and KSC, where MALNISI achieves substantially larger radar coverage in comparison to alternative similarity measures. For PaviaU, MALNISI provides measurable improvements across all the metrics ($\sim 4$\% improvement in NMI). In the HongHu dataset, all similarity measures, except Mahalanobis, which achieves the highest AA, attain nearly identical results. For the Salinas dataset, all similarity measures yield nearly identical performance. This behavior is primarily due to its high spectral separability and large homogeneous regions ($2$-nd column of Table~\ref{tab:visual_results}), which diminish the need for specialized graph-weighting strategies. In Botswana, SAM and Dice outperform MALNISI on certain metrics, implying that scale-invariant angular similarity and structural overlap measure better align with the sparse characteristics of this dataset. The comparatively weaker performance of Minkowski, SID, SSIM, and SIFT across several datasets further highlights the importance of selecting a graph metric suitable for HSI.
\begin{figure}[ht]
    \centering
    \includegraphics[width=\linewidth]{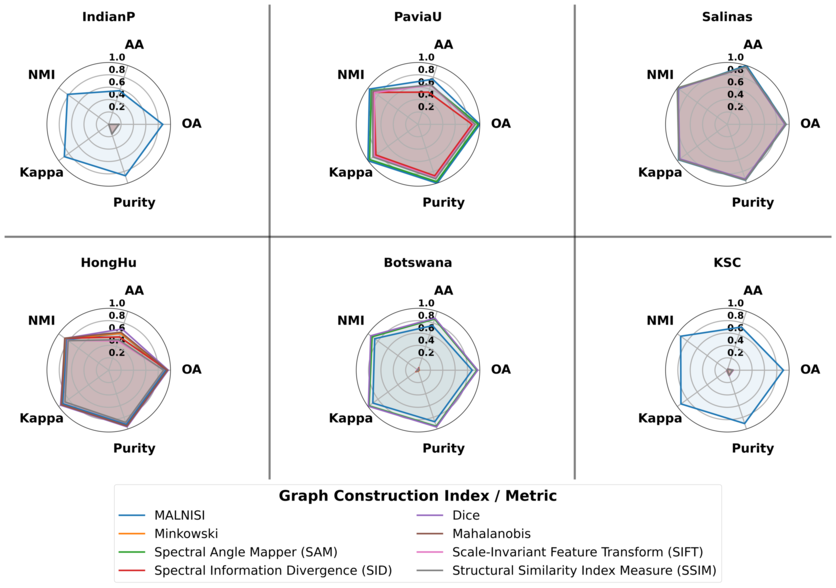}
    \caption{Radar visualization of performance analysis of different similarity indices for segmentation. The angular axes represent different evaluation metrics, the radial axes reflect their normalized range of values, and each polygon corresponds to the effectiveness of the similarity measure.}
    \label{fig:metric_radar}
\end{figure}

\textbf{D. Impact of different loss functions: }
The contribution of the proposed Weighted Contrastive Loss (WCL) is examined against standard Contrastive Loss (CL), Reconstruction Loss, Triplet Loss, Laplacian Total Variation Loss, and Orthogonal Loss. The improvement on using WCL is evident in IndianP, PaviaU and HongHu, where it achieves larger radar coverage across all metrics (except AA in IndianP). Standard CL exhibits comparatively lower performance, indicating that unweighted positive pair treatment may overlook local neighborhood reliability and introduce weaker embedding consistency. Other losses perform worse, suggesting that preserving reconstruction fidelity or enforcing pairwise margins alone is insufficient to capture complex nonlinear manifold structures in HSI data. For Salinas, all methods achieve highly comparable performance, with only marginal differences across evaluation metrics. This suggests that strong intrinsic spectral separability reduces sensitivity to the choice of loss function, allowing even simpler objectives to learn sufficiently discriminative embeddings. In Botswana and KSC, the performance gap between WCL and standard CL becomes relatively smaller, although WCL remains competitive or marginally superior. This behavior indicates that sparse and compact latent structures reduce the dependency on heavily guided contrastive weighting. Nevertheless, the consistent advantage of WCL demonstrates that MALNISI-guided weighting contributes positively across varying HSI characteristics.
\begin{figure}[ht]
    \centering
    \includegraphics[width=\linewidth]{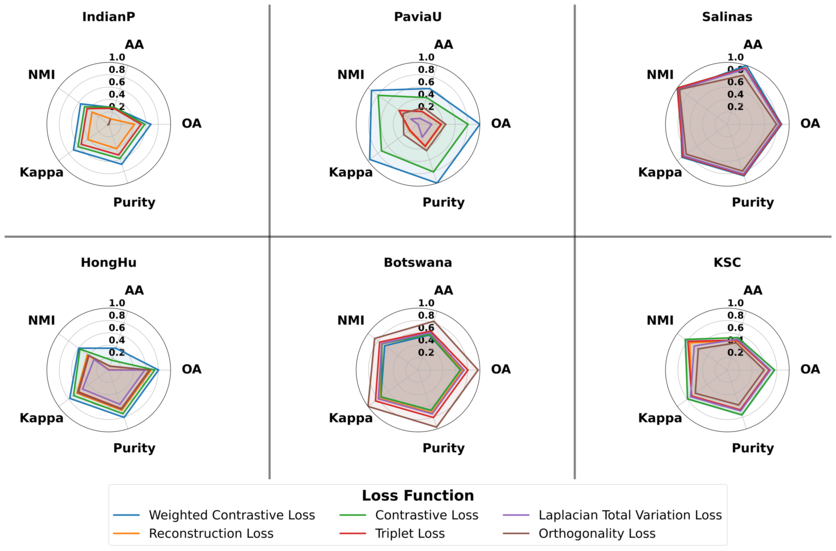}
    \caption{Radar visualization of performance analysis of different loss functions.}
    \label{fig:loss_radar}
\end{figure}

\begin{figure}[ht]
    \centering
    \includegraphics[width=\linewidth]{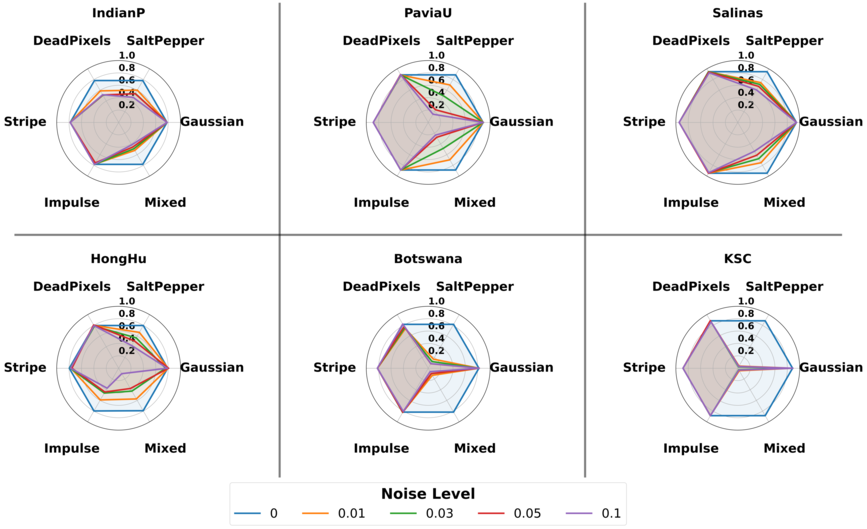}
    \caption{Radar visualization of performance analysis under different noise perturbations. The angular axes represent different noise types, the radial axes represent normalized NMI values, and each polygon corresponds to different noise perturbations.}
    \label{fig:noise_radar}
\end{figure}

\textbf{E. Noise Sensitivity Analysis: }
Comparative radar visualization (Figure \ref{fig:noise_radar}) of NMI under $6$ noise types \cite{gonzalez2009digital}: Gaussian, Salt and Pepper, Dead Pixels, Stripe, Impulse, and Mixed shows the proposed method's lower noise sensitivity except on Salt and Pepper and Mixed. Salt and Pepper and Mixed noise produce the steepest NMI degradation, particularly for PaviaU, Botswana, and KSC at higher corruption levels ($0.05$--$0.1$). Salt and Pepper introduce many random extreme-valued perturbations distributed throughout the image, severely disturbing neighborhood consistency during graph construction, and as MALNISI relies on local neighborhood similarity, these noise levels lead to widespread corruptions and misleading adaptive weights. Mixed noise exhibits the strongest deterioration because it combines multiple corruption mechanisms. The degradation in dead pixels and impulse noise in certain cases, in spite of the noise being isolated, may locally affect MALNISI weighting, as they primarily disturb only a limited set of nodes whose values either become $0$ or are degraded, respectively. Thus, rather than globally disrupting graph connectivity, the surrounding clean neighbors dominate, graph topology is largely preserved, and graph attention aggregation is able to further suppress unreliable isolated nodes, resulting in smaller performance degradation compared to salt and pepper noise.

\begin{figure}[ht]
    \centering
    \includegraphics[width=\linewidth]{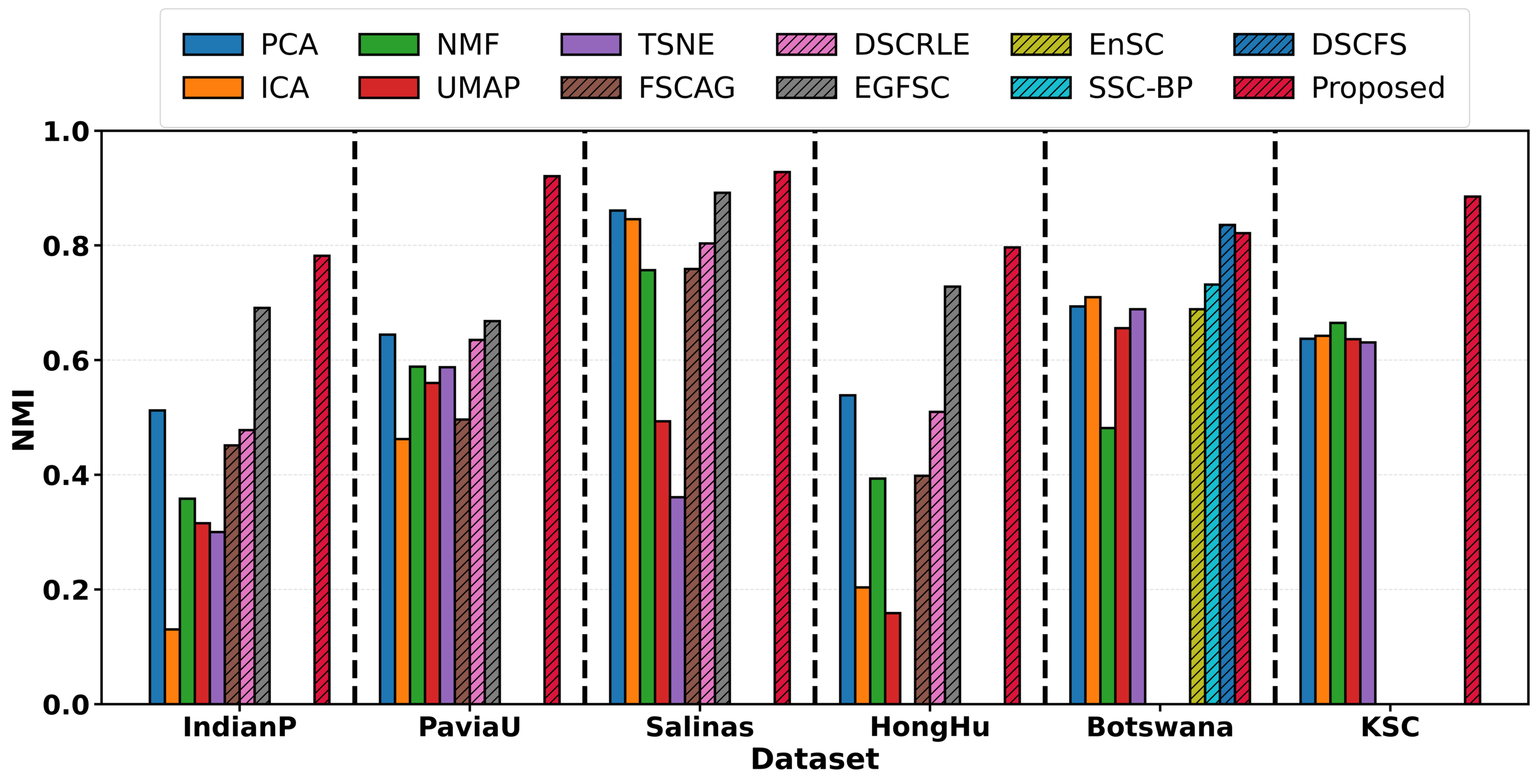}
    \caption{NMI comparison of traditional and modern feature embedding methods on $6$ benchmark HSI datasets. Solid bars denote traditional feature extraction techniques, while hatched bars denote modern embedding methods.}
    \label{fig:feature_embedding}
\end{figure}

\textbf{F. Comparative Analysis of Unsupervised Traditional and Modern Feature Extraction Methods: }
Figure \ref{fig:feature_embedding} compares the clustering performance of traditional and modern unsupervised embedding methods on $6$ benchmark HSI datasets using NMI as the evaluation metric. Among the traditional feature extraction methods, including PCA, ICA, NMF, UMAP, and t-SNE, performance varies considerably across datasets. The proposed framework substantially outperforms all traditional methods on every dataset. A similar trend is observed when comparing against modern graph- and subspace-based embedding methods on all datasets except Botswana, where the performance is comparable to the strongest baseline, DSCFS. It is worth noting that not all modern methods report results on every benchmark dataset; therefore, some bars are absent in the figure. Overall, the proposed framework consistently achieves the best or highly competitive performance across all datasets, demonstrating its ability to learn discriminative representations that are well-suited for HSI clustering.


The performance impact of different clustering algorithms and the influence of spatial neighborhood size in MALNISI are presented in Section \ref{app:results} of the supplementary documents. The analysis shows the superiority of SC and the non-significant influence of neighborhood size on the performance.

\section{Conclusion \& Scope For Future Works} \label{CFW}
The proposed framework does unsupervised segmentation of both well-separated/complex, compact/non-compact, and non-linear HSI data with improved performance. It accurately groups spatially distributed regions by simultaneously increasing inter-LULC discrimination and decreasing intra-LULC variations during latent space embedding. Extensive experiments validate the specific effectiveness of the three proposed modules in achieving this enhanced separability. The architecture also demonstrates strong robustness against spectral noise interference. Besides, the model consistently outperforms traditional and modern unsupervised feature extractors, which typically fail to capture the complex nonlinearities of hyperspectral data. Finally, the introduction of MALNISI provides significantly more reliable and context-aware graph weights compared to conventional spectral similarity indices, establishing a solid structural foundation for the network.

Despite its strong performance, the proposed framework has some limitations. The learned representations depend on the quality of the constructed graph, making them sensitive to inaccurate neighborhood relationships. Moreover, graph construction requires predefined neighborhood parameters that may vary across datasets. The graph construction, weighted contrastive loss, and spectral clustering are computationally expensive for large hyperspectral scenes. Furthermore, the proposed framework does not explicitly address sparse data, limiting its effectiveness on datasets with sparse spatial distributions. Most hyperspectral image datasets also exhibit severe class imbalance, which degrades the clustering performance of minority classes. Future work will focus on developing mechanisms to effectively handle sparse data and class imbalance while enabling adaptive end-to-end graph learning with more robust graph-aware contrastive objectives and self-supervised learning to better exploit spectral–spatial information. Scalability can be improved through efficient graph sparsification, approximate spectral clustering, or hierarchical graph partitioning techniques. Finally, dynamic neighborhood modelling also offers promising directions for further advancing large-scale hyperspectral image clustering.

\section*{Acknowledgments}
{The authors would like to thank the Indian Statistical Institute, Kolkata, for their institutional support. The authors are also grateful to their colleagues and all those who provided valuable feedback and assistance during the course of this research.}

\bibliographystyle{IEEEtran}
\bibliography{references}

\section*{Biography}
\begin{IEEEbiography}[{\includegraphics[width=1in,height=1.25in,clip,keepaspectratio]{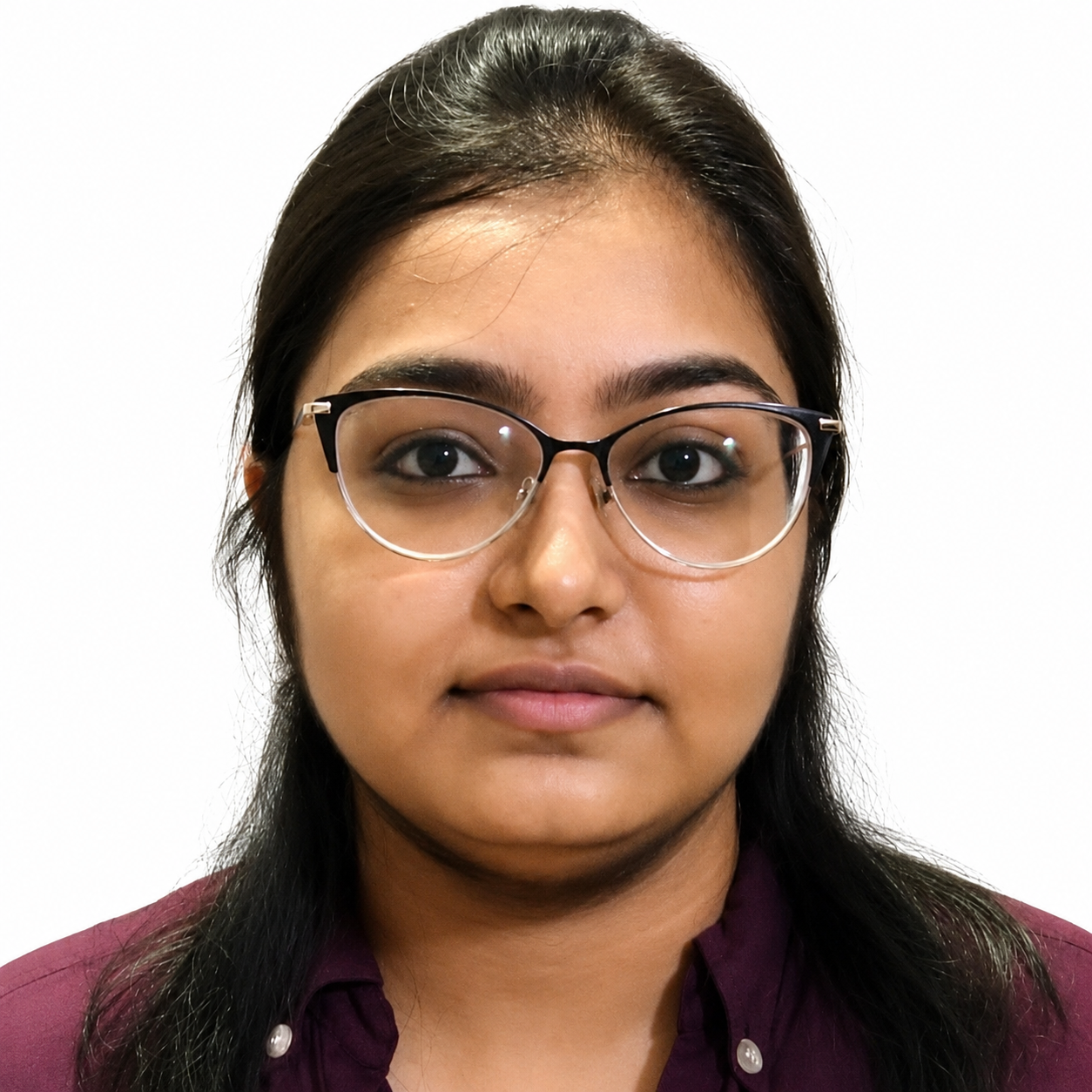}}]{Soumi Sarkar}
is a research scholar at the Machine Intelligence Unit in the Indian Statistical Institute, Kolkata. She has completed her M.Tech. in CSE and M.Sc. in CIS from the University of Calcutta in Kolkata and her B.Sc. in CS from Bethune College, affiliated with Calcutta University, in Kolkata. She is currently working on the application of deep learning methods in remote sensing applications.
\end{IEEEbiography}
\begin{IEEEbiography}[{\includegraphics[width=1in,height=1.25in,clip,keepaspectratio]{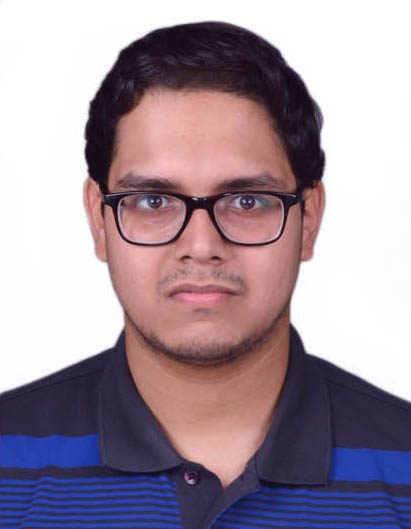}}]{Subhadip Boral}
received the Ph.D. in Computer Science from University of Calcutta in 2026. He is currently working as an Assistant Professor in the Computer Science \& Information Technology, Siksha 'O' Anusandhan, Bhubaneswar, Odisha.
\end{IEEEbiography}
\begin{IEEEbiography}[{\includegraphics[width=1in,height=1.25in,clip,keepaspectratio]{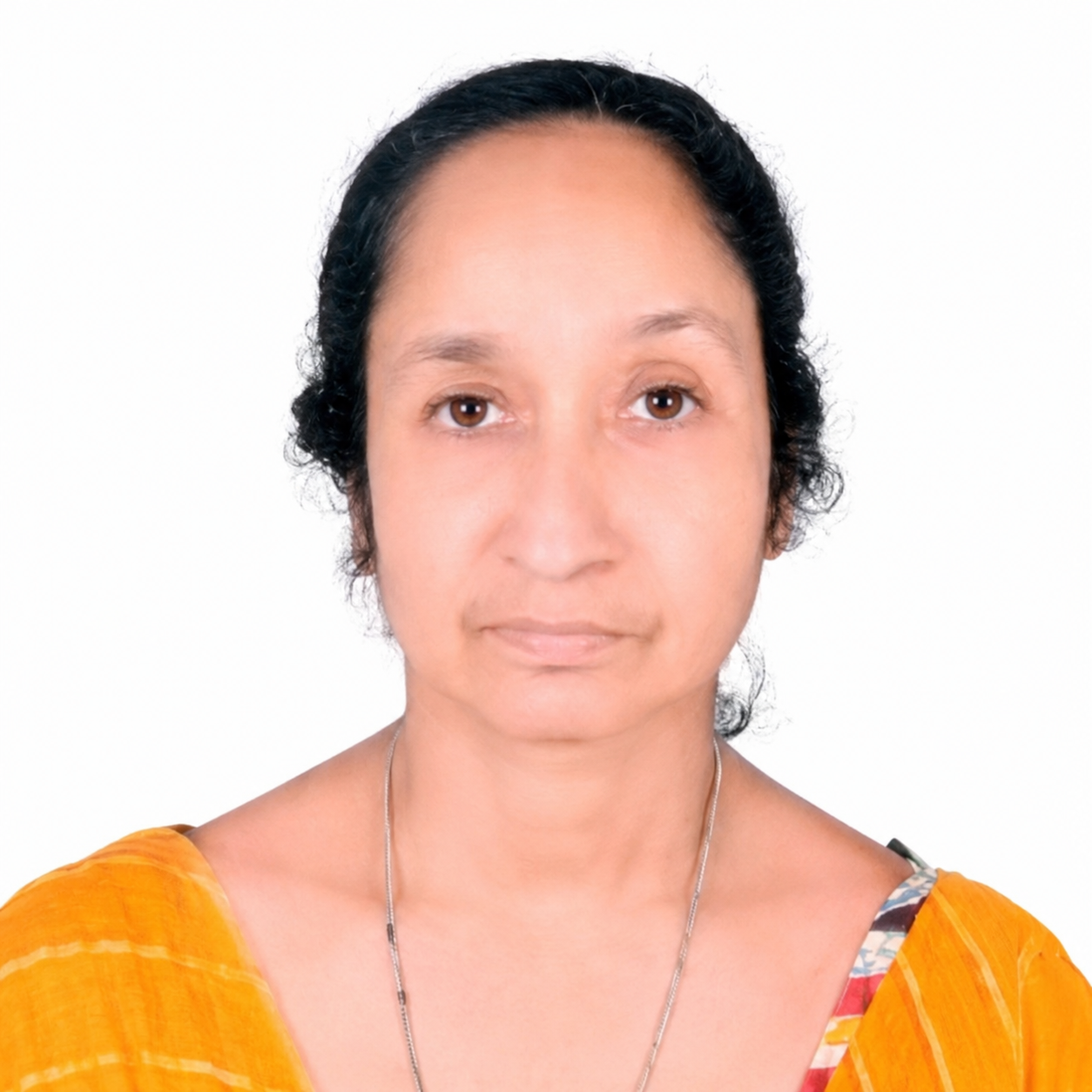}}]{Sarbani Palit}
received her B.Tech (Hons) from the Indian Institute of Technology, Kharagpur, India, in 1987 and her M.S. degree from the University of California, Santa Barbara, in 1989 and her Ph.D. from the Indian Institute of Technology, Kharagpur, India, in 1994. She has more than twenty-three years of academic experience. She is currently working as a faculty member in Computer Vision and Pattern Recognition Unit, Indian Statistical Institute, India.
\end{IEEEbiography}
\begin{IEEEbiography}[{\includegraphics[width=1in,height=1.25in,clip,keepaspectratio]{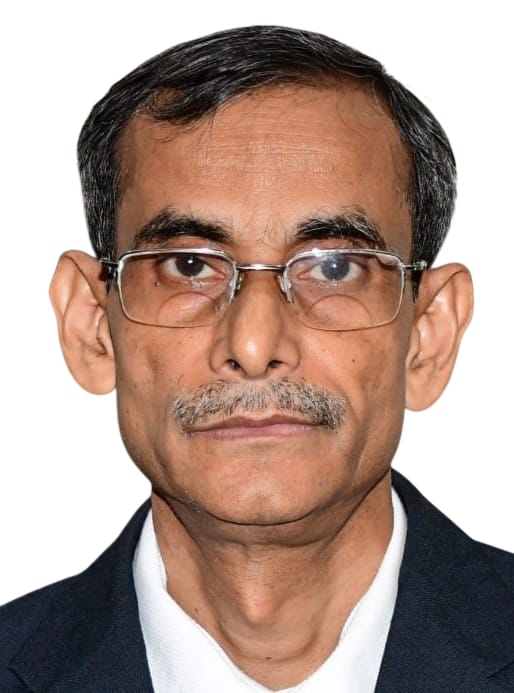}}]{Ashish Ghosh}
is the Director of the International Institute of Information Technology, Bhubaneswar, Odisha. He is also a Senior Professor at the Indian Statistical Institute. He has published more than 300 research papers in international journals and conferences, and acting as the Principal Investigator of several funded projects. His current research interests include Machine and Deep Learning, Data Science, Image/Video Analysis, and Computational Intelligence.
 \end{IEEEbiography}

\end{document}